\pdfoutput=1
\documentclass[11pt]{article}

\usepackage[final]{acl}

\usepackage{times}
\usepackage{latexsym}
\usepackage[colorinlistoftodos]{todonotes}

\usepackage[T1]{fontenc}

\usepackage[utf8]{inputenc}

\usepackage{microtype}

\usepackage{inconsolata}

\usepackage{graphicx}

\usepackage{dsfont}

\usepackage{enumitem}
\usepackage{rotating}

\usepackage{booktabs}
\usepackage{multirow}
\usepackage{tcolorbox}

\usepackage{subcaption}

\usepackage{arydshln}

\usepackage{longtable}
\usepackage{booktabs}

\usepackage{multirow}
\usepackage{makecell}
\usepackage{amsthm}
\usepackage{amsmath}
\usepackage{amssymb}
\usepackage{xspace}
\usepackage{cleveref}
\usepackage{minted}
\usepackage{algorithm}
\usepackage{algpseudocode}
\usepackage{threeparttable}
\usepackage{enumitem}
\newcommand{\tpl}[1]{\texttt{<#1>}}
\newcommand{\narrowtextsc}[1]{\textls[-50]{\textsc{#1}}}

\newcommand{\sys}[1]{\narrowtextsc{#1}}
\newcommand{\data}[1]{\textsf{\small#1}}
\newcommand{\datatab}[1]{\textsf{\tiny#1}}

\usepackage{xcolor,colortbl}
\usepackage[dvipsnames]{xcolor}

\usepackage{soul}
\definecolor{lightblue}{RGB}{239,247,250}
\definecolor{blue}{RGB}{224,230,255}
\definecolor{lighpurple}{RGB}{226, 218, 246}
\definecolor{purple}{RGB}{193, 175, 236}
\definecolor{deepred}{RGB}{183,26,26}
\definecolor{deepgreen}{RGB}{4,98,10}
\definecolor{lightred}{RGB}{242,207, 194}
\definecolor{LightGreen}{RGB}{230,250,245}
\definecolor{LightOrange}{RGB}{255,240,235}
\definecolor{colbach}{rgb}{0.3, 0.3, 1}

\usepackage{verbatim}

\usepackage{tcolorbox}
\tcbuselibrary{skins,breakable}

\usepackage{listings}          
\usepackage{xcolor}            

\lstdefinestyle{predictionbox}{
    basicstyle=\ttfamily\small,
    backgroundcolor=\color{gray!8},
    frame=single,
    rulecolor=\color{gray!40},
    framerule=0.6pt,
    xleftmargin=2mm,
    xrightmargin=2mm,
    breaklines=true,           
    breakatwhitespace=true,
    tabsize=4,
    columns=fullflexible,      
    keepspaces=true,
    aboveskip=6pt,
    belowskip=6pt
}

\author{
    Van Bach Nguyen\textsuperscript{1} \qquad
    Qianli Wang\textsuperscript{2,4} \qquad
    \\
    \textbf{Nils Feldhus}\textsuperscript{3,4,\footnotemark[2]} \qquad
    \textbf{Jörg Schlötterer}\textsuperscript{1} \qquad
    \textbf{Christin Seifert}\textsuperscript{1}
    \\
    \\
    \textsuperscript{1} University of Marburg \quad
    \textsuperscript{2} Technische Universit\"at Berlin \quad
    \textsuperscript{3} University of Groningen \\
    \textsuperscript{4} German Research Center for Artificial Intelligence (DFKI) \\
}

\title{\sys{SimXAI}: A Simple Drop-in Layer Increases Interaction Quality and Understanding in Conversational XAI}

\begin{document}
\maketitle

\renewcommand{\thefootnote}{\fnsymbol{footnote}}
\footnotetext[2]{Work done at TU Berlin.}
\renewcommand*{\thefootnote}{\arabic{footnote}}

\begin{abstract}
Conversational explainable AI (ConvXAI) agents often use GUS-style architectures with separate modules for intent recognition, dialogue management, and template-based response generation. These pipelines are hard to adapt and often struggle with context-dependent follow-up questions.
We present \sys{SimXAI}, a lightweight LLM-centric conversational layer that replaces these modules with a single tool-calling LLM that bridges access to existing XAI resources. Rather than a standalone agent, \sys{SimXAI} serves as a modular interface that enables conversational access to external XAI resources. We further introduce an evaluation suite for single- and multi-turn parsing, response-to-tool faithfulness, and contextualization. Across existing and new datasets, \sys{SimXAI} achieves strong function-selection accuracy and high response-to-tool faithfulness while better handling context-dependent requests. In a user study replacing only the conversational layer of \sys{TalkToModel}, a representative GUS-based agent, users completed tasks more accurately and in fewer turns. Overall, \sys{SimXAI} improves conversational interaction while leaving the underlying XAI resources unchanged\footnote{The source code (including evaluation suite) is available at: \url{https://github.com/bach-mr/SimXAI}}.
\end{abstract}

\section{Introduction}
\begin{figure}[t]
    \centering
    \includegraphics[width=\linewidth]{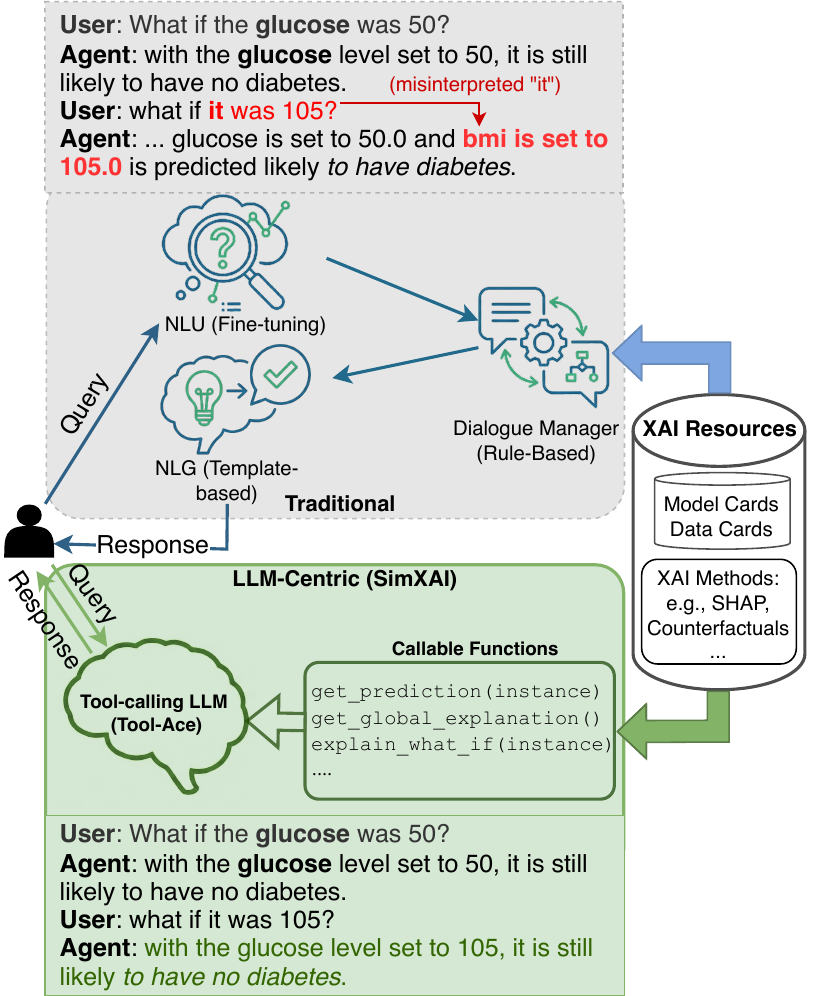}
    \caption{Comparison of \sys{TalkToModel}'s traditional GUS architecture with \sys{SimXAI}'s LLM-centric conversational component. Traditional systems rely on multiple components that require rules, templates, or fine-tuning. In contrast, \sys{SimXAI} requires only function specifications and a single LLM. Both architectures share the same XAI resources. }
    \label{fig:intro}
\end{figure}
The XAI community has reached a consensus that explanations should take a conversational form~\cite{miller_explanation_2019, lakkaraju2022rethinkingexplainabilitydialoguepractitioners, zhang-etal-2025-convxai}. Conversational explainable AI (ConvXAI) agents therefore move beyond one-off explanations and embrace conversational approaches to provide explanations. These conversational explanations have been demonstrated to be more effective for users' model understanding \cite{shen-2023-convxai, mindlin2024measuring, he-etal-2025-convxai}. 
Conceptually, a ConvXAI agent can be decomposed into two parts: (i) a conversational layer (\textbf{ConvXAI layer}), responsible for managing the interaction with the user, and (ii) the underlying \textbf{XAI resources}, which store or generate the explanatory information (e.g., model/data cards~\cite{mitchell2019model, pushkarna2022data},  or feature-attribution methods like SHAP~\cite{lundberg2017unified} or counterfactual explanations~\cite{mothilal2020explaining,ross2021explaining}). Most current ConvXAI agents (i.a., \citealt{nguyen2023black, slack-2023-talktomodel, feldhus-etal-2023-interrolang, wang-etal-2024-llmcheckup}) adopt the traditional GUS (Genial Understanding System) architecture~\cite{bobrow1977} for the ConvXAI layer, which comprises three distinct components: \textit{natural language understanding (NLU)} for intent recognition, \textit{dialogue management} for context maintenance and rule application, and \textit{natural language generation (NLG)} for response production.
Although such GUS-based ConvXAI layers present explanations through interactive dialogue, they rely on \textbf{multiple models separately fine-tuned} on \textbf{handcrafted  data} for intent recognition in NLU, \textbf{rule-based} dialogue management, and \textbf{template-based} response generation in NLG (Figure~\ref{fig:intro}). As a result, they remain limited in contextualization, scalability, and generalization~\cite{Rastogi_Zang_Sunkara_Gupta_Khaitan_2020}. These limitations highlight the need for lightweight, adaptive, and flexible ConvXAI layers. Recent developments suggest that LLMs can fulfill this role and replace traditional ConvXAI layers in such systems~\cite{li-etal-2024-large-language-models}. 

\begin{table*}[ht!]
\centering
\renewcommand*{\arraystretch}{1}
\resizebox{\textwidth}{!}{
\begin{tabular}{ccccc}
\toprule
\textbf{System} & \textbf{Contextualization} &  \textbf{Scalability} & \textbf{Generalization} & \shortstack{\textbf{Response}\\\textbf{faithfulness}}\\
\midrule
\sys{TalkToModel} \cite{slack-2023-talktomodel} & Medium  & Low (FT, D, T, R)  & Low  & High \\
\sys{InterroLang} \cite{feldhus-etal-2023-interrolang} & Weak  & Low (FT, D, T, R)  & Low  & High \\
\sys{XAgent}~\cite{nguyen2023black}    & Weak  & Medium (FT, D, T, R)  &  Low & High \\

\textbf{\sys{SimXAI}} & \textbf{Strong} & \textbf{High} & \textbf{High} & \textbf{High} \\

\bottomrule
\end{tabular}
}
\caption{Design-level comparison of \sys{SimXAI} with the conversational layers of prior GUS-based ConvXAI agents in terms of contextualization, scalability, generalization, and response-to-resource faithfulness. The table summarizes architectural properties from the respective system designs. Scalability constraints include: Finetuning (FT), Handcrafted Data (D), Templates (T), and Rules (R).}
\label{tab:xai-comparison}
\end{table*}

We introduce \textbf{\sys{SimXAI}}, a simple, lightweight, LLM-centric framework that acts as a ConvXAI layer between users and existing XAI resources (see~\Cref{fig:intro}). Unlike rule- or template-based systems, \sys{SimXAI} uses an LLM to support natural, context-aware dialogue without requiring handcrafted intent datasets or conversational fine-tuning. The layer is \textbf{data- and model-agnostic}. Integrating it into an agent requires only function implementations that provide access to the agent's existing XAI resources. \sys{SimXAI} can therefore replace the conversational components of GUS-based agents (e.g., \citealt{slack-2023-talktomodel,nguyen2023black}) while preserving the underlying explanations. \textbf{\sys{SimXAI} is not a standalone agent} and does not itself select or validate XAI methods. Instead, it provides a modular interface to resources supplied by the host system.

Using LLMs for tool calling is not technically novel. Our contribution is a ConvXAI-specific design that separates a replaceable conversational layer from XAI resources. The design also provides a function interface based on XAI question types and an evaluation framework that separately measures function parsing, response-to-tool faithfulness, and contextualization.

Prior work on evaluating ConvXAI has mainly focused on isolated turns~\cite{malandri2023convxai, wang-etal-2024-llmcheckup} and has often overlooked dialogue context. To address this gap, we evaluate \sys{SimXAI} on parsing accuracy in single- and multi-turn dialogues, contextualization, and response-to-tool faithfulness. For each aspect, we provide a dataset with manually annotated ground truth, corresponding metrics, and an automated evaluation pipeline. The faithfulness evaluation measures whether the conversational layer preserves the returned XAI output. It does not measure whether the underlying explanation method is faithful to the predictive model.


In a user study with 30 participants, replacing the ConvXAI layer of \sys{TalkToModel} with \sys{SimXAI} while keeping the underlying XAI resources unchanged, significantly improved both, perceived interaction quality and task performance.

In summary, our contributions are threefold. \textbf{First}, based on the analysis of limitations of prior work (\S\ref{sec:limitations}), we introduce \sys{SimXAI}, an LLM-centric framework that addresses these limitations and offers a practical alternative to conventional ConvXAI layers (\S\ref{sec:simxai}). \textbf{Second}, we present a multi-aspect evaluation suite for ConvXAI layers, including automated assessment of parsing, response-to-tool faithfulness, and contextualization (\S\ref{sec:evaluation}). \textbf{Third}, we show through quantitative evaluation and a user study that replacing a conventional conversational layer with \sys{SimXAI} improves significantly conversational quality while preserving faithfulness to underlying XAI outputs (\S\ref{sec:results}, \S\ref{sec:comparison}).

\section{Limitations of Conversational Layers in Prior ConvXAI Agents}
\label{sec:limitations}

GUS-based ConvXAI layers (such as \citealt{nguyen2023black, slack-2023-talktomodel, feldhus-etal-2023-interrolang}) support follow-up questions, but their conversational behavior is typically implemented through task-specific grammars, intent schemas, templates, or dialogue rules. As summarized in Table~\ref{tab:xai-comparison}, this design leads to a recurring trade-off between \textit{faithfulness}  and three other dimensions: \textit{contextualization}, \textit{scalability}, and \textit{generalization}.

First, prior systems provide only \textit{weak-to-moderate contextualization}, handling context primarily through pre-defined operational structures rather than flexible discourse-level reasoning. \sys{TalkToModel} resolves follow-up questions through grammar operators such as \texttt{previous\_filter} and \texttt{previous\_operation}~\cite{slack-2023-talktomodel}. \sys{InterroLang} extends this design with intent recognition, slot filling, reference resolution, and clarification strategies~\cite{feldhus-etal-2023-interrolang}. \sys{XAgent} supports multi-turn interaction through a fixed dialogue manager and a curated phrase bank~\cite{nguyen2023black}. Such mechanisms can handle follow-up questions only to a limited extent and often struggle when users rely on ellipsis, anaphora, or long-range dependencies (see~\Cref{fig:intro} for an example of a failure case). \looseness=-1

Second, this design leads to \textit{low-to-moderate scalability} and \textit{limited generalization} across domains and tasks. \sys{TalkToModel} requires dataset-specific grammars and synthetic utterance--parse pairs~\cite{slack-2023-talktomodel}, \sys{InterroLang} depends on annotated intents, slots, and task-specific operations~\cite{feldhus-etal-2023-interrolang}, and \sys{XAgent} relies on manually curated phrases and response templates~\cite{nguyen2023black}. As a result, extending such systems to new settings usually requires additional annotation or rule engineering rather than simply integrating new XAI resources. This also limits generalization: prior systems are typically demonstrated in narrow settings, such as tabular data or NLP-specific pipelines, and their conversational control logic does not transfer easily beyond those conditions. 

At the same time, these constrained designs maximally preserve \textit{faithfulness} in response generation. As they generate outputs through rules or templates, the conversational layer is less likely to introduce unsupported claims or misrepresent the underlying XAI method results. The problem with prior systems is not faithfulness itself, but the rigid control logic they use to maintain it. That logic limits how well prior systems handle context and makes these systems hard to adapt across domains. 

In contrast, \sys{SimXAI} addresses these problems by replacing handcrafted conversational control logic with a tool-calling LLM that bridges access to existing XAI resources, without requiring finetuning, handcrafted data, or templates. This design enables \textit{strong contextualization}, \textit{high scalability} and \textit{high generalization}, as it does not rely on domain-specific implementations.

\section{SimXAI}
\label{sec:simxai}

SimXAI adds a conversational interface to existing XAI resources using only two components: (1) an off-the-shelf LLM with tool-calling support and (2) a set of general function specifications that map user requests onto available XAI resources. The novelty lies in this ConvXAI-specific separation and interface, rather than in the general mechanism of LLM tool calling. Unlike prior systems, \sys{SimXAI} does not rely on task-specific grammars, intent schemas, dialogue managers, or response templates. Instead, \sys{SimXAI} interprets the current user query together with the dialogue history, selects the appropriate function, fills its arguments from context when applicable, and verbalizes the returned result. The backend remains responsible for computing the prediction, explanation, or data/model information returned by that function.

Formally, given a user query $q$, dialogue history $H$, and a set of available functions $\mathcal{T}$, the system either answers directly or invokes a function $t \in \mathcal{T}$. When a function call is required, the system extracts the necessary arguments from $q$ and $H$, asks for clarification if required information cannot be recovered from the dialogue context, executes the function, and generates a final response grounded in the returned output. This design replaces handcrafted conversational control logic with a single tool-calling LLM interface that bridges access to existing XAI resources. Algorithm~\ref{alg:tool_augmented_generation} summarizes this interaction loop.
\begin{algorithm}[t]
\caption{Response Generation}
\label{alg:tool_augmented_generation}
\small
\begin{algorithmic}[1]
\Require User query $q$, dialogue history $H$, tool set $\mathcal{T}$
\Ensure Final response $r$

\State $u \gets \textsc{DecideToolUsage}(q, H, \mathcal{T})$
\If{$u = \text{None}$}
    \State $r \gets \textsc{InternalKnowledge}(q, H)$
    \State \Return $r$
\EndIf

\State $t \gets \textsc{SelectTool}(q, H, \mathcal{T})$
\State $a \gets \textsc{ExtractArguments}(t, q, H)$

\While{$\neg \textsc{Valid}(a, t)$}
    \State $m \gets \textsc{FindMissingArguments}(a, t)$
    \State $a' \gets \textsc{SearchInDialogue}(m, H)$
    \If{$a' \neq \emptyset$}
        \State $a \gets \textsc{UpdateArguments}(a, a')$
    \Else
        \State $c \gets \textsc{AskUserForClarification}(m)$
        \State $H \gets H \cup \{c\}$
        \State $a \gets \textsc{UpdateArguments}(a, c)$
    \EndIf
\EndWhile

\State $o \gets \textsc{CallTool}(t, a)$
\State $r \gets \textsc{GenerateResponse}(q, H, o)$
\State \Return $r$

\end{algorithmic}
\end{algorithm}

\subsection{Tool-Calling LLM}
\label{subsec:model}
We use ToolACE-2.5-Llama-3.1-8B\footnote{\url{https://huggingface.co/Team-ACE/ToolACE-2.5-Llama-3.1-8B}}, an off-the-shelf LLM trained for tool-use reasoning~\cite{liu2025toolace,grattafiori2024llama3herdmodels}. In \sys{SimXAI}, the LLM is responsible for four decisions: (i) whether tool use is needed, (ii) if so, which function to call, (iii) how to ground its arguments in the current turn and dialogue history, and (iv) how to generate a final natural-language response from the function output. No additional parser, dialogue manager, or template bank is required.

\begin{table*}[t]
\footnotesize
\centering
\begin{tabular}{p{7cm} p{8cm}}
\toprule
\textbf{Function} & \textbf{Description} \\
\midrule
\texttt{get\_prediction(instance)} & Return the model prediction for an instance. \\
\texttt{get\_global\_explanation()} & Provide a global explanation of model behavior. \\
\texttt{get\_local\_explanation(instance, prediction\_label)} & Explain why the model produced a specific prediction. \\
\texttt{explain\_why\_not(instance, expected\_label)} & Explain why an expected label was not predicted. \\
\texttt{explain\_how\_to\_be\_that(instance, target\_label)} & Suggest how the input could be changed to obtain a target label. \\
\texttt{explain\_how\_to\_still\_be\_this(instance)} & Identify changes that would preserve the current prediction. \\
\texttt{explain\_what\_if(instance)} & Show how predictions change under input modifications. \\
\texttt{get\_model\_performance(metric)} & Report a model performance metric. \\
\texttt{get\_data\_information()} & Provide information about the training and evaluation data. \\
\texttt{get\_model\_information()} & Describe the model, training setup, strengths, and limitations. \\
\texttt{get\_output\_information()} & Explain the meaning of the model outputs. \\
\bottomrule
\end{tabular}
\caption{Core functions used by \sys{SimXAI}.}
\label{tab:functions}
\end{table*}
\subsection{Function Set}
Our set of functions is based on the question bank of \citet{liao2022humancenteredexplainableaixai, liao2020questioning}. This question bank was created from practitioner interviews and organized around model-, data-, and context-related questions about AI systems. From this bank, we implement 11 functions: one retrieves the model prediction for a given instance and the other ten address explanatory questions, such as \emph{why} the model made a particular prediction or \emph{how} to modify an instance to change the prediction. For each function, we specify its purpose, inputs, arguments, and outputs. \Cref{tab:functions} summarizes the full set. The set is extensible: if no predefined function matches a user’s intent, a new function can be added with its own specification (\Cref{fig:operation-template}).

\paragraph{Function Specification.}
\label{subsec: operation_description}
Function specification is critical for ensuring that the LLM calls the correct function. For each function, we adopt a standard specification template (\Cref{fig:operation-template}) that defines the function name, argument list and types, documentation, and expected input and output types. \textit{These specifications are not tied to any particular dataset or model}. Instead, we use general specifications that apply across datasets and models, enabling broad generalization (cf. \Cref{fig:function-example} in the appendix for a concrete example of a counterfactual function specification).
\begin{figure}[h]
\centering
\begin{lstlisting}[style=predictionbox]
function_name(argument: type) -> return_type
    # summary of what the function does.
    Args:
        argument (type): Description of the       input (format, constraints).
    Returns:
        return_type: Description of the           output (labels, score range, etc.).
    
\end{lstlisting}
\caption{Template for Function Specification.}
\label{fig:operation-template}
\end{figure}

\section{Evaluation}
\label{sec:evaluation}

We evaluate ConvXAI layers along three aspects: \textit{parsing}, \textit{response-to-tool faithfulness}, and \textit{contextualization}. Parsing measures whether the system selects the correct function and arguments for a user request, response-to-tool faithfulness measures whether the final response is supported by the returned XAI output, and contextualization measures whether the system uses dialogue history appropriately when interpreting the current turn. Table~\ref{tab:eval_suite} summarizes the evaluation suite. This scope evaluates the conversational layer only: it neither measures nor guarantees the intrinsic correctness or model-faithfulness of the backend XAI method that produced the tool output.
We use this suite for two purposes. First, we assess \sys{SimXAI} directly across settings of increasing conversational complexity. Second, we qualitatively analyze gains from replacing the conversational components of prior GUS-based ConvXAI systems with \sys{SimXAI} while keeping their underlying XAI resources unchanged.

\subsection{Parsing}
\label{subsec:Parsing}
Parsing evaluates whether the system maps a user request to the correct function and arguments. We report exact-match accuracy over the selected function. Unlike prior work, which mainly evaluates parsing in single-turn settings~\cite{slack-2023-talktomodel,feldhus-etal-2023-interrolang,nguyen2023black}, we consider three levels of conversational complexity.

\paragraph{Single-turn parsing.}
We evaluate standard single-turn function selection on two datasets in ConvXAI settings: \data{CoXQL}~\cite{wang-etal-2024-coxql} for explanation request parsing (Figure~\ref{fig:coxql}) and \data{Compass}~\cite{wang-etal-2025-multilingual-datasets} for user input extraction (Figure~\ref{fig:parsing}). In this setting, the current query can be interpreted without dialogue history.

\paragraph{Multi-turn parsing.}
To evaluate short-range contextual parsing, we construct \data{Compass-MultiTurn} from \data{Compass} by merging related single-turn interactions into short conversations and removing redundant user turns (details in Appendix~\ref{sec:parsing}). In this setting, the use of recent dialogue history is required to parse the users' intent correctly.

\paragraph{Dialogue-level parsing.}
To evaluate long-range dependencies, we generate synthetic dialogue-level datasets in two domains: sentiment analysis \data{SYN-Sentiment} and heart rate prediction \data{SYN-HeartRate}. Each of the two datasets combines dialogues generated under two settings: \textbf{Constrained}, in which we provide our 11 function specifications in the prompt and ask GPT-5.1 to generate realistic conversations covering all operations, and \textbf{Unconstrained}, in which GPT-5.1 is prompted to generate realistic conversations without any predefined constraints. Each user turn defines one parsing instance, with the full preceding dialogue provided as context. 
Human validation assessed these synthetic dialogues as mostly realistic (details in Appendix~\ref{apdx:sec:user_study_realism}).

Together, these datasets assess whether ConvXAI systems can select the correct function corresponding to user's intent. They range from isolated questions up to full multi-turn conversations with long range dependencies. Table~\ref{tab:test_set} summarizes the parsing datasets.
\begin{table}[h]
\centering
\small
\resizebox{\linewidth}{!}{
\begin{tabular}{lccc}
\toprule
\textbf{Dataset} & \textbf{\#Dialogues} & \textbf{\#Turns} & \textbf{\#Queries} \\
\midrule
\data{CoXQL} + \data{Compass} & -- & 1 & 462 \\
\data{Compass-MultiTurn} & 10 & 4--6 & 26 \\
\data{SYN-Sentiment} & 4 & 12--24 & 40 \\
\data{SYN-HeartRate} & 4 & 12--24 & 31 \\
\bottomrule
\end{tabular}
}
\caption{Statistics for parsing datasets with increasing conversational complexity.}
\label{tab:test_set}
\end{table}

\subsection{Response-to-Tool Faithfulness}

As LLMs are prone to hallucination~\cite{huang2025survey}, we evaluate the faithfulness of the agent's responses with respect to the provided XAI resources. Following prior work~\cite{adlakha-etal-2024-evaluating}, we define \emph{response-to-tool faithfulness} as the extent to which the conversational layer preserves the information contained in a returned XAI output without introducing unsupported claims. This is distinct from the intrinsic faithfulness of the XAI method to the predictive model, which is outside our scope.

We construct two faithfulness datasets, \data{SYN-QA-Sentiment} and \data{SYN-QA-HeartRate}, based on \data{SYN-Sentiment} and \data{SYN-HeartRate}. The datasets contain 71 question–answer pairs covering prediction lookup, explanation outputs, model descriptions, and dataset information. For each question, we manually construct a gold standard function output and evaluate whether the system response faithfully preserves this information using a binary judgment and accuracy as aggregated metric.


\subsection{Contextualization}

Contextualization evaluates whether the system uses dialogue history appropriately to answer the current query. We first assess \emph{context relevance} by manually determining whether a query can be answered independently or requires the preceding dialogue. For queries that depend on prior context, we then assess whether the response incorporates the necessary dialogue history to resolve the current turn coherently. This \textbf{\emph{contextual answer coherence}} is a binary judgment, aggregated as accuracy (the fraction of contextually coherent answers).

\begin{table*}[t]
\centering
\scriptsize
\resizebox{\textwidth}{!}{
\begin{tabular}{lp{1.15cm}p{5.2cm}lr}
\toprule
\textbf{Aspect} & \textbf{Subtask} & \textbf{What is evaluated} & \textbf{Dataset} & \textbf{\sys{SimXAI}} \\
\midrule
\multirow{4}{*}{Parsing}
& Single-turn
& Correct function selection from a single user query
& \datatab{CoXQL}, \datatab{Compass}
& 86\% \\

& Multi-turn
& Correct function selection from short dialogue history
& \datatab{Compass-MultiTurn}
& 88\%\\

& \multirow{2}{1.15cm}{Dialogue-level}
& \multirow{2}{5.2cm}{Correct function selection from full conversations with long-range dependencies}
& \datatab{SYN-Sentiment} 
& 92.5\%\\

& 
& 
& \datatab{SYN-HeartRate} 
& 96\%\\
\midrule

\multirow{2}{*}{Faithfulness}
& --
& \multirow{2}{4.2cm}{Whether the response is fully supported by the underlying XAI output}
& \datatab{SYN-QA-Sentiment}
& 93.6\%\\

& --
& 
& \datatab{SYN-QA-HeartRate}
& 91.5\%\\
\midrule

\multirow{4}{*}{Contextualization}
& \multirow{2}{1.15cm}{Contextual relevance}
& \multirow{2}{4.2cm}{Whether the current query requires prior dialogue context}
& \datatab{SYN-Sentiment} 
& N/A\\

& 
& 
& \datatab{SYN-HeartRate}
& N/A\\
& \multirow{2}{1.15cm}{Contextual coherence}
& \multirow{2}{4.2cm}{Whether responses to context-dependent queries use prior context coherently}
& \datatab{SYN-Sentiment} 
& 90.9\%\\

& 
& 
& \datatab{SYN-HeartRate}
& 94.74\%\\
\bottomrule
\end{tabular}
}
\caption{Overview of the evaluation suite for the ConvXAI layer and the corresponding accuracy results of \sys{SimXAI}.}
\label{tab:eval_suite}
\end{table*}
\section{Results}
\label{sec:results}
To assess \sys{SimXAI}, we report performance on parsing, response-to-tool faithfulness, and contextual coherence.
Following recent work on LLM-based evaluation~\cite{xu-etal-2025-context, tamber-etal-2025-benchmarking}, we additionally examine opportunities for automated evaluation by measuring the agreement between human rating and LLM-as-a-judge.\footnote{We use \url{https://huggingface.co/meta-llama/Llama-3.1-8B-Instruct}}
Specifically, we report how often the LLM judge assigns the same label as the human annotator for response-to-tool faithfulness and contextualization, including \textit{context relevance} and \textit{contextual coherence}.

\subsection{SimXAI Performance}
\Cref{tab:eval_suite} reports results across the three aspects in the evaluation suite: parsing, response-to-tool faithfulness, and contextualization.

For \emph{parsing}, \sys{SimXAI} performs strongly in all settings, achieving above 86\% accuracy for single-turn, multi-turn, and dialogue-level evaluation. Interestingly, accuracy is higher on the multi-turn and dialogue-level datasets, suggesting that dialogue context helps the model to resolve user intents better.

In most cases (over 91\%), \sys{SimXAI} produces responses that are \textit{faithful to the tool output}, indicating that the conversational layer preserves the returned XAI information rather than introducing unsupported claims. This is particularly important because \sys{SimXAI} replaces rule- and template-based conversational components with an LLM-based interface, which may increase the risk of hallucinations. These results do not evaluate the intrinsic faithfulness of the backend XAI methods.

\sys{SimXAI} further shows high \textit{contextual answer coherence }across domains: 90.91\% for sentiment analysis and 94.74\% for heart rate, indicating that prior dialogue context is generally incorporated correctly in context-dependent queries.

\textit{Overall, \sys{SimXAI} performs well across all three aspects of the evaluation suite. Parsing accuracy remains high from single-turn to dialogue-level evaluation, responses are usually supported by the returned XAI outputs, and most context-dependent queries are answered coherently.}

\subsection{LLM-as-a-judge Performance}
\begin{figure}[t!]
\centering
\resizebox{\columnwidth}{!}{
\begin{minipage}{\columnwidth}
\includegraphics[width=\columnwidth]{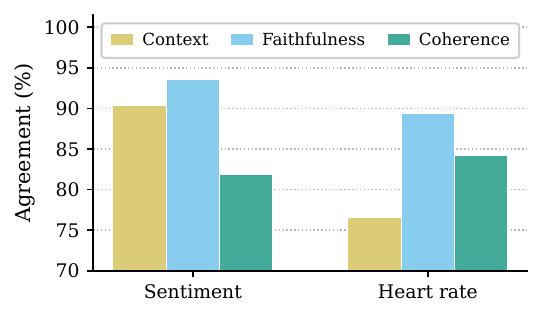}
\end{minipage}
}
\caption{Agreement between LLM-as-a-judge and human annotation.}
\label{fig:llm-judge-results}
\end{figure}

\Cref{fig:llm-judge-results} shows the agreement between LLM-as-a-judge and human annotation. The agreement on \textbf{response-to-tool faithfulness} is high (around 90\%), suggesting that LLM-based evaluation is a promising proxy for this aspect on our datasets.
While the agreement on \emph{contextual relevance} in sentiment analysis is equally high (around 90\%), it drops to 76.60\% in heart rate prediction. This indicates that automatically determining whether a query depends on prior context remains challenging in some domains.
For \emph{contextual answer coherence}, the LLM judge agrees with human annotations in more than 80\% of cases across both domains, suggesting that it can serve as a useful proxy for coherence evaluation.

\textit{In summary, evaluation with LLM-as-a-judge shows potential for automated assessment of response-to-tool faithfulness and contextualization. Despite lower agreement on context relevance in heart rate prediction, it may still serve as an efficient approach for initial comparative evaluation.}




\section{GUS-based system comparison}
\label{sec:comparison}
\sys{SimXAI} is designed to be integrated into existing conversational XAI agents by replacing their original conversational components. We qualitatively examine how \sys{SimXAI} can improve existing systems, by replacing the conversational layer of the ConvXAI agents \sys{TalkToModel}~\cite{slack-2023-talktomodel}, \sys{InterroLang}~\cite{feldhus-etal-2023-interrolang}, and \sys{XAgent}~\cite{nguyen2023black} with \sys{SimXAI} and re-assess cases, in which the original system failed (e.g., due to missing context handling abilities).

For each baseline, we keep the original function set and underlying XAI outputs fixed, and replace only the module that maps user requests and dialogue history onto function selection and response generation. This isolates the contribution of \sys{SimXAI} acting solely as conversational layer and not as a modified ConvXAI backend.

We further conduct a user study on the integration of \sys{SimXAI} into \sys{TalkToModel}, a representative grammar-based system. 
We directly compare the original model with its \sys{SimXAI}-integrated  counterpart in terms of both, objective and subjective metrics. 

\subsection{Qualitative results}
\Cref{fig:intro} shows an example failure case of incorrect context handling of \sys{TalkToModel}'s native conversational layers.
\sys{TalkToModel} fails to maintain contextual understanding because the agent misinterprets the referent of ``it'' (anaphora), which leads to an incorrect final decision. Replacing \sys{TalkToModel}'s conversational layers by \sys{SimXAI} fixes this issue and correctly resolves the context in the same conversation dialogue. We observe similar patterns of incorrect context handling in the other agents under comparison, which are partially unable to provide an answer at all. Replacing the conversational layers of these agents by \sys{SimXAI} corrects such issues (examples in Appendix~\ref{apdx:sec:qualitative_examples}).

\subsection{User Study}
We conduct a controlled within-subjects study to evaluate whether integrating \sys{SimXAI} into \sys{TalkToModel} \cite{slack-2023-talktomodel} 1) improves the perceived quality of the conversation and 2) leads to a better understanding of the machine learning model. To isolate the effect of the conversational component, we keep all other components (datasets, classifiers, and XAI resources) unchanged. We compare two agents: (i) the original \sys{TalkToModel} with its fine-tuned ConvXAI layers and (ii) \sys{TalkToModel} with \sys{SimXAI} replacing the ConvXAI layers. Both agents follow the original \sys{TalkToModel} configuration described by \citet{slack-2023-talktomodel}.

\subsubsection{Setup}
We recruited 30 lay participants via Prolific.\footnote{\url{https://www.prolific.com/} Participants receive a compensation of \$8/hour} This sample size aligns with prior HCI and XAI studies evaluating interactive explanation systems \cite{kaur2020interpreting,slack-2023-talktomodel}. Participants interact with both agents using the Diabetes dataset~\cite{diabetes_34}, as in the original \sys{TalkToModel} study.

For each agent, participants complete three tasks. These tasks cover multiple categories from an established XAI question taxonomy\cite{liao2020questioning}, including data understanding, local explanations, and counterfactual inspection. Each task contains at least two questions, for a total of nine questions across all tasks. The questions are sampled from a pool of predefined multiple-choice questions for each task (details in Appendix \Cref{apdx:subsec:question_pool}). They are designed to be answerable using the ground-truth information provided by the agent’s explanations. Some questions by design require multi-turn interactions to arrive at the correct answer, in order to assess conversational coherence and context handling. We randomize the order of agents, and although the question types remain consistent across conditions, specific values (e.g., instance identifiers) are varied to mitigate learning effects.

We collect both, objective and subjective measures. \textbf{Objective metrics} are (1) answer accuracy, (2) conversational efficiency measured as the number of turns per correctly answered question, which is commonly used in dialogue agent evaluation \cite{su-etal-2016-line,peng-etal-2017-composite}, and (3) indicators of conversational breakdowns, such as repeated or rephrased questions and agent failures, where the agent does not provide meaningful answers. \textbf{Subjective measures} consist of two parts. First, we use the 5-point Likert BUS-11 usability scale~\cite{borsci2024re}, which includes 11 items across five subscales, focusing on perceived conversation and information quality and perceived chatbot functionality. Second, we include a \textbf{pairwise comparison} questionnaire between the two agents to assess user preference and explanation effectiveness.

\subsubsection{Results}
This section summarizes the key objective and subjective results. Details on the significance tests for the reported metrics, the full BUS-11 scale, and the complete pairwise comparison are provided in Appendix \Cref{apdx:subsec:detail_result}.
\begin{table}[t]
\centering
\small
\resizebox{\linewidth}{!}{
\begin{tabular}{lcc}
\toprule
\textbf{Measure} & \textbf{\sys{TalkToModel}} & \textbf{\sys{SimXAI}} \\
\midrule
Accuracy $\uparrow$ & 59.0 & \textbf{92.1} \\
Chat turns $\downarrow$ & 12.4 & \textbf{10.0} \\
Chat time (min) $\downarrow$ & 7.4 & \textbf{5.8} \\
Turns per correct answer $\downarrow$ & 2.70 & \textbf{1.32} \\

\bottomrule
\end{tabular}
}
\caption{Objective results from the user study.}
\label{tab:user_study_objective}
\end{table}

\begin{table}[t]
\centering
\small
\resizebox{\linewidth}{!}{
\begin{tabular}{lcc}
\toprule
\textbf{BUS-11 item (shortened)} & \textbf{\sys{TalkToModel}} & \textbf{\sys{SimXAI}} \\
\midrule
Interaction Clarity & 3.33  & \textbf{4.60}  \\
Context Awareness & 3.13  & \textbf{4.37}  \\
Response Comprehensibility & 3.60  & \textbf{4.57}  \\
Goal Facilitation & 3.10  & \textbf{4.37}  \\
Information Sufficiency & 3.23  & \textbf{4.37}  \\
Information Relevance & 3.13  & \textbf{4.13}  \\
Response Accuracy & 3.63  & \textbf{4.47}  \\
Response Time & \textbf{4.37}  & 4.03  \\
\midrule
Cronbach's $\alpha$ & 0.925 & 0.856  \\
\bottomrule
\end{tabular}
}
\caption{Mean BUS-11 ratings from the user study (1 = strongly disagree, 5 = strongly agree). We report eight relevant items in shortened form, full results are in \Cref{tab:userstudy_bus11_means}. Cronbach's $\alpha > 0.8$ indicates high internal consistency.}
\label{tab:user_study_bus11}
\end{table}
\paragraph{Objective results.} Replacing \sys{TalkToModel}’s conversational component with \sys{SimXAI} leads to clear improvements in both, task performance and perceived interaction quality. \Cref{tab:user_study_objective} shows that participants achieve substantially higher accuracy with \sys{SimXAI} than with the original \sys{TalkToModel} (92.1\% vs.\ 59.0\%, $p<.001$). \sys{SimXAI} also reduces interaction costs: users complete the task in fewer turns on average (10.0 vs.\ 12.4 turns) and in less time (5.8 vs.\ 7.4 minutes). Across all sessions, this translates to roughly half as many turns per correct answer with \sys{SimXAI} (1.32 vs.\ 2.70).

\paragraph{Subjective results.} Table~\ref{tab:user_study_bus11} reports the BUS-11 usability ratings. We observe the largest improvements in clarity of communication, context tracking, helpfulness in achieving the user’s goal, and providing an appropriate amount of information (all $p<.001$). 
Perceived response time is the only item where \sys{SimXAI} scores lower, suggesting a possible trade-off between conversational quality and latency. We attribute this to the models deployed for parsing in \sys{TalkToModel}, which are comparably smaller (ranging from 60M to 6B) than to the tool-augmented LLM we use in \sys{SimXAI} (\S\ref{subsec:model}). However, the difference is not statistically significant (p>0.1).

\paragraph{Pairwise comparison.} 80.0\% of the participants prefer \sys{SimXAI} overall, while 16.7\% prefer \sys{TalkToModel} and 3.3\% indicate no preference (cf. \Cref{tab:userstudy_preference} in the Appendix). We observe the same proportions for perceived task effectiveness. 


\textit{Overall, the user study shows that replacing the conversational component of \sys{TalkToModel} with \sys{SimXAI} improves both,  perceived interaction quality and task performance, even when the underlying XAI resources remain unchanged.}
\section{Related Work}
Besides the systems analyzed in \S\ref{sec:limitations}, that rely on fine-tuned models, off-the-shelf LLMs are increasingly adopted in ConvXAI. \Citet{van2024xagent} extend \sys{XAgent} by Llama-2~\cite{touvron2023llama2openfoundation} for postprocessing NLG and answering general factual questions (e.g., about machine learning terminology).
\sys{KnowThySelf}~\cite{prasai2025knowthyselfagenticassistantllm} is a multi-agent framework, which at its core still resembles the GUS architecture. It employs Gemma3-27B~\cite{gemmateam2025gemma3technicalreport} for most parts of NLU and NLG, while user intents are matched to specialized agents by embedding-based similarity, using nomic-embed-text~\cite{nussbaum2025nomic}. These specialized agents encapsulate XAI resources. 
Closest to our work is \sys{Echo}~\cite{sebe-etal-2025-echo}, another agentic framework with a tool-calling LLM at its core. However, \sys{Echo} trades faithfulness for human validation ``... might not always result in a helpful
or correct answer [...] This further highlights the importance of human-in-the-loop workflows, in which the LLM can try again until the answer is satisfactory''. This undermines the purpose of ConvXAI to explain a model's true reasoning, as any plausible answer is likely satisfactory. Yet, plausibility and correctness are different evaluation criteria that should not be conflated~\cite{10.1145/3583558} (a plausible answer might not be correct, while a correct one might not be plausible, e.g., if the model's reasoning is flawed).

All these systems are demos without quantitative or qualitative evaluation.
Unlike prior GUS-based systems or existing tool-calling ConvXAI agents, our contribution is not the generic tool-calling mechanism itself. \sys{SimXAI} introduces a simple, lightweight, open-source intermediate layer with a ConvXAI-specific separation between conversational control and underlying XAI resources, an extensible XAI function design, and a dedicated evaluation suite. The measured faithfulness in this suite is response-to-tool faithfulness and should not be interpreted as validation of the backend explanation method.

\section{Conclusion}
\sys{SimXAI} enables natural conversational interactions with existing XAI resources through general callable functions. Across evaluations of parsing, response-to-tool faithfulness, and contextualization, its lightweight LLM-centric design supports conversational explanations while usually preserving the information returned by the XAI resources.
Our user study shows that replacing the conversational layers of a traditional GUS-based ConvXAI agent by \sys{SimXAI} significantly improves task performance and perceived interaction quality. This suggests that LLM-centric architectures provide tangible benefits to users in understanding AI model behavior.
In future work, we plan to explore and enhance fully automated evaluation for LLM-based conversational XAI layers. Based on that, we plan to study how LLM choice affects tool calling and to develop further improvements to ConvXAI.

\section*{Limitations}
Like other LLM-based systems, \sys{SimXAI} introduces additional latency and remains susceptible to hallucinations. 

\sys{SimXAI} improves only the conversational component: it assumes that appropriate XAI resources are already available and does not assess the intrinsic quality, correctness, or model-faithfulness of the underlying explanation methods themselves. Consequently, a response can be faithful to a tool output even when that output is incomplete or misleading. 

We do not study the impact of the LLM choice for tool calling or LLM-as-a-judge. Our experiments use a lightweight 8B LLaMA 3.1 model, and results may vary with other models.


\bibliography{custom}

@inproceedings{feldhus-etal-2023-interrolang,
    title = "{I}nterro{L}ang: Exploring {NLP} Models and Datasets through Dialogue-based Explanations",
    author = {Feldhus, Nils  and
      Wang, Qianli  and
      Anikina, Tatiana  and
      Chopra, Sahil  and
      Oguz, Cennet  and
      M{\"o}ller, Sebastian},
    editor = "Bouamor, Houda  and
      Pino, Juan  and
      Bali, Kalika",
    booktitle = "Findings of the Association for Computational Linguistics: EMNLP 2023",
    month = dec,
    year = "2023",
    address = "Singapore",
    publisher = "Association for Computational Linguistics",
    url = "https://aclanthology.org/2023.findings-emnlp.359",
    doi = "10.18653/v1/2023.findings-emnlp.359",
    pages = "5399--5421"
}

@inproceedings{he-etal-2025-convxai,
    author = {He, Gaole and Aishwarya, Nilay and Gadiraju, Ujwal},
    title = {Is Conversational XAI All You Need? Human-AI Decision Making With a Conversational XAI Assistant},
    year = {2025},
    isbn = {9798400713064},
    publisher = {Association for Computing Machinery},
    address = {New York, NY, USA},
    url = {https://doi.org/10.1145/3708359.3712133},
    doi = {10.1145/3708359.3712133},
    booktitle = {Proceedings of the 30th International Conference on Intelligent User Interfaces},
    pages = {907–924},
    numpages = {18},
    location = {
    },
    series = {IUI '25}
}

@misc{lakkaraju2022rethinkingexplainabilitydialoguepractitioners,
      title={Rethinking Explainability as a Dialogue: A Practitioner's Perspective}, 
      author={Himabindu Lakkaraju and Dylan Slack and Yuxin Chen and Chenhao Tan and Sameer Singh},
      year={2022},
      eprint={2202.01875},
      archivePrefix={arXiv},
      primaryClass={cs.LG},
      url={https://arxiv.org/abs/2202.01875}, 
}

@article{malandri2023convxai,
  title={Convxai: a system for multimodal interaction with any black-box explainer},
  author={Malandri, Lorenzo and Mercorio, Fabio and Mezzanzanica, Mario and Nobani, Navid},
  journal={Cognitive Computation},
  volume={15},
  number={2},
  pages={613--644},
  year={2023},
  publisher={Springer}
}

@incollection{mindlin2024measuring,
  title={Measuring User Understanding in Dialogue-Based xAI Systems},
  author={Mindlin, Dimitry and Robrecht, Amelie Sophie and Morasch, Michael and Cimiano, Philipp},
  booktitle={ECAI 2024},
  pages={1148--1155},
  year={2024},
  publisher={IOS Press}
}

@misc{prasai2025knowthyselfagenticassistantllm,
      title={KnowThyself: An Agentic Assistant for LLM Interpretability}, 
      author={Suraj Prasai and Mengnan Du and Ying Zhang and Fan Yang},
      year={2025},
      eprint={2511.03878},
      archivePrefix={arXiv},
      primaryClass={cs.AI},
      url={https://arxiv.org/abs/2511.03878},
}

@article{sebe-etal-2025-echo,
    author = {Vanbrabant, Sebe and Eerlings, Gilles and Rovelo Ruiz, Gustavo Alberto and Vanacken, Davy},
    title = {ECHO: Enhancing Conversational Explainable AI through Tool-Augmented Language Models},
    year = {2025},
    issue_date = {June 2025},
    publisher = {Association for Computing Machinery},
    address = {New York, NY, USA},
    volume = {9},
    number = {4},
    url = {https://doi.org/10.1145/3734191},
    doi = {10.1145/3734191},
    journal = {Proc. ACM Hum.-Comput. Interact.},
    month = jun,
    articleno = {EICS014},
    numpages = {33}
}

@inproceedings{shen-2023-convxai,
	author = {Shen, Hua and Huang, Chieh-Yang and Wu, Tongshuang and Huang, Ting-Hao Kenneth},
	title = {Conv{XAI}: Delivering Heterogeneous {AI} Explanations via Conversations to Support Human-{AI} Scientific Writing},
	year = {2023},
	isbn = {9798400701290},
	publisher = {Association for Computing Machinery},
	address = {New York, NY, USA},
	url = {https://doi.org/10.1145/3584931.3607492},
	doi = {10.1145/3584931.3607492},
	booktitle = {Computer Supported Cooperative Work and Social Computing},
	pages = {384–387},
	numpages = {4},
	location = {Minneapolis, MN, USA},
	series = {CSCW '23 Companion}
}

@article{slack-2023-talktomodel,
	author = {Slack, Dylan
		and Krishna, Satyapriya
		and Lakkaraju, Himabindu
		and Singh, Sameer},
	title = {Explaining machine learning models with interactive natural language conversations using {T}alk{T}o{M}odel},
	journal = {Nature Machine Intelligence},
	year = {2023},
	month = {Jul},
	day = {27},
	issn = {2522-5839},
	doi = {10.1038/s42256-023-00692-8},
	url = {https://doi.org/10.1038/s42256-023-00692-8}
}

@inproceedings{wang-etal-2024-llmcheckup,
    title = "{LLMC}heckup: Conversational Examination of Large Language Models via Interpretability Tools and Self-Explanations",
    author = {Wang, Qianli  and
      Anikina, Tatiana  and
      Feldhus, Nils  and
      Genabith, Josef  and
      Hennig, Leonhard  and
      M{\"o}ller, Sebastian},
    editor = "Blodgett, Su Lin  and
      Curry, Amanda Cercas  and
      Dey, Sunipa  and
      Madaio, Michael  and
      Nenkova, Ani  and
      Yang, Diyi  and
      Xiao, Ziang",
    booktitle = "Proceedings of the Third Workshop on Bridging Human--Computer Interaction and Natural Language Processing",
    month = jun,
    year = "2024",
    address = "Mexico City, Mexico",
    publisher = "Association for Computational Linguistics",
    url = "https://aclanthology.org/2024.hcinlp-1.9",
    doi = "10.18653/v1/2024.hcinlp-1.9",
    pages = "89--104",
}

@inproceedings{wang-etal-2024-coxql,
    title = "{C}o{XQL}: A Dataset for Parsing Explanation Requests in Conversational {XAI} Systems",
    author = {Wang, Qianli  and
      Anikina, Tatiana  and
      Feldhus, Nils  and
      Ostermann, Simon  and
      M{\"o}ller, Sebastian},
    editor = "Al-Onaizan, Yaser  and
      Bansal, Mohit  and
      Chen, Yun-Nung",
    booktitle = "Findings of the Association for Computational Linguistics: EMNLP 2024",
    month = nov,
    year = "2024",
    address = "Miami, Florida, USA",
    publisher = "Association for Computational Linguistics",
    url = "https://aclanthology.org/2024.findings-emnlp.76/",
    doi = "10.18653/v1/2024.findings-emnlp.76",
    pages = "1410--1422",
}

@inproceedings{wang-etal-2025-multilingual-datasets,
    title = "Multilingual Datasets for Custom Input Extraction and Explanation Requests Parsing in Conversational {XAI} Systems",
    author = {Wang, Qianli  and
      Anikina, Tatiana  and
      Feldhus, Nils  and
      Ostermann, Simon  and
      Splitt, Fedor  and
      Li, Jiaao  and
      Tsoneva, Yoana  and
      M{\"o}ller, Sebastian  and
      Schmitt, Vera},
    editor = "Christodoulopoulos, Christos  and
      Chakraborty, Tanmoy  and
      Rose, Carolyn  and
      Peng, Violet",
    booktitle = "Findings of the Association for Computational Linguistics: EMNLP 2025",
    month = nov,
    year = "2025",
    address = "Suzhou, China",
    publisher = "Association for Computational Linguistics",
    url = "https://aclanthology.org/2025.findings-emnlp.29/",
    pages = "534--555",
    ISBN = "979-8-89176-335-7",
}

@inproceedings{zhang-etal-2025-convxai,
    author = {Zhang, Tong and Zhang, Mengao and Low, Wei Yan and Yang, X. Jessie and Li, Boyang Albert},
    title = {Conversational Explanations: Discussing Explainable AI with Non-AI Experts},
    year = {2025},
    isbn = {9798400713064},
    publisher = {Association for Computing Machinery},
    address = {New York, NY, USA},
    url = {https://doi.org/10.1145/3708359.3712143},
    doi = {10.1145/3708359.3712143},
    booktitle = {Proceedings of the 30th International Conference on Intelligent User Interfaces},
    pages = {409–424},
    numpages = {16},
    location = {
    },
    series = {IUI '25}
}

@inproceedings{liu2025toolace,
 author = {Liu, Weiwen and Huang, Xu and Zeng, Xingshan and hao, xinlong and Yu, Shuai and Li, Dexun and Wang, Shuai and Gan, Weinan and Liu, Zhengying and Yu, Yuanqing and WANG, Zezhong and Wang, Yuxian and Ning, Wu and Hou, Yutai and Wang, Bin and Wu, Chuhan and Xinzhi, Wang and Liu, Yong and Wang, Yasheng and Tang, Duyu and Tu, Dandan and Shang, Lifeng and Jiang, Xin and Tang, Ruiming and Lian, Defu and Liu, Qun and Chen, Enhong},
 booktitle = {International Conference on Representation Learning},
 editor = {Y. Yue and A. Garg and N. Peng and F. Sha and R. Yu},
 pages = {41359--41381},
 title = {ToolACE: Winning the Points of LLM Function Calling},
 url = {https://proceedings.iclr.cc/paper_files/paper/2025/file/663865ea167425c6c562cb0b6bcf76c7-Paper-Conference.pdf},
 volume = {2025},
 year = {2025}
}

@inproceedings{nguyen2023black,
  title={From black boxes to conversations: Incorporating xai in a conversational agent},
  author={Nguyen, Van Bach and Schl{\"o}tterer, J{\"o}rg and Seifert, Christin},
  booktitle={World Conference on Explainable Artificial Intelligence},
  pages={71--96},
  year={2023},
  organization={Springer}
}

@misc{grattafiori2024llama3herdmodels,
      title={The Llama 3 Herd of Models}, 
      author={Aaron Grattafiori and Abhimanyu Dubey and Abhinav Jauhri and Abhinav Pandey and Abhishek Kadian and Ahmad Al-Dahle and Aiesha Letman and Akhil Mathur and Alan Schelten and Alex Vaughan and Amy Yang and Angela Fan and Anirudh Goyal and Anthony Hartshorn and Aobo Yang and Archi Mitra and Archie Sravankumar and Artem Korenev and Arthur Hinsvark and Arun Rao and Aston Zhang and Aurelien Rodriguez and Austen Gregerson and Ava Spataru and Baptiste Roziere and Bethany Biron and Binh Tang and Bobbie Chern and Charlotte Caucheteux and Chaya Nayak and Chloe Bi and Chris Marra and Chris McConnell and Christian Keller and Christophe Touret and Chunyang Wu and Corinne Wong and Cristian Canton Ferrer and Cyrus Nikolaidis and Damien Allonsius and Daniel Song and Danielle Pintz and Danny Livshits and Danny Wyatt and David Esiobu and Dhruv Choudhary and Dhruv Mahajan and Diego Garcia-Olano and Diego Perino and Dieuwke Hupkes and Egor Lakomkin and Ehab AlBadawy and Elina Lobanova and Emily Dinan and Eric Michael Smith and Filip Radenovic and Francisco Guzmán and Frank Zhang and Gabriel Synnaeve and Gabrielle Lee and Georgia Lewis Anderson and Govind Thattai and Graeme Nail and Gregoire Mialon and Guan Pang and Guillem Cucurell and Hailey Nguyen and Hannah Korevaar and Hu Xu and Hugo Touvron and Iliyan Zarov and Imanol Arrieta Ibarra and Isabel Kloumann and Ishan Misra and Ivan Evtimov and Jack Zhang and Jade Copet and Jaewon Lee and Jan Geffert and Jana Vranes and Jason Park and Jay Mahadeokar and Jeet Shah and Jelmer van der Linde and Jennifer Billock and Jenny Hong and Jenya Lee and Jeremy Fu and Jianfeng Chi and Jianyu Huang and Jiawen Liu and Jie Wang and Jiecao Yu and Joanna Bitton and Joe Spisak and Jongsoo Park and Joseph Rocca and Joshua Johnstun and Joshua Saxe and Junteng Jia and Kalyan Vasuden Alwala and Karthik Prasad and Kartikeya Upasani and Kate Plawiak and Ke Li and Kenneth Heafield and Kevin Stone and Khalid El-Arini and Krithika Iyer and Kshitiz Malik and Kuenley Chiu and Kunal Bhalla and Kushal Lakhotia and Lauren Rantala-Yeary and Laurens van der Maaten and Lawrence Chen and Liang Tan and Liz Jenkins and Louis Martin and Lovish Madaan and Lubo Malo and Lukas Blecher and Lukas Landzaat and Luke de Oliveira and Madeline Muzzi and Mahesh Pasupuleti and Mannat Singh and Manohar Paluri and Marcin Kardas and Maria Tsimpoukelli and Mathew Oldham and Mathieu Rita and Maya Pavlova and Melanie Kambadur and Mike Lewis and Min Si and Mitesh Kumar Singh and Mona Hassan and Naman Goyal and Narjes Torabi and Nikolay Bashlykov and Nikolay Bogoychev and Niladri Chatterji and Ning Zhang and Olivier Duchenne and Onur Çelebi and Patrick Alrassy and Pengchuan Zhang and Pengwei Li and Petar Vasic and Peter Weng and Prajjwal Bhargava and Pratik Dubal and Praveen Krishnan and Punit Singh Koura and Puxin Xu and Qing He and Qingxiao Dong and Ragavan Srinivasan and Raj Ganapathy and Ramon Calderer and Ricardo Silveira Cabral and Robert Stojnic and Roberta Raileanu and Rohan Maheswari and Rohit Girdhar and Rohit Patel and Romain Sauvestre and Ronnie Polidoro and Roshan Sumbaly and Ross Taylor and Ruan Silva and Rui Hou and Rui Wang and Saghar Hosseini and Sahana Chennabasappa and Sanjay Singh and Sean Bell and Seohyun Sonia Kim and Sergey Edunov and Shaoliang Nie and Sharan Narang and Sharath Raparthy and Sheng Shen and Shengye Wan and Shruti Bhosale and Shun Zhang and Simon Vandenhende and Soumya Batra and Spencer Whitman and Sten Sootla and Stephane Collot and Suchin Gururangan and Sydney Borodinsky and Tamar Herman and Tara Fowler and Tarek Sheasha and Thomas Georgiou and Thomas Scialom and Tobias Speckbacher and Todor Mihaylov and Tong Xiao and Ujjwal Karn and Vedanuj Goswami and Vibhor Gupta and Vignesh Ramanathan and Viktor Kerkez and Vincent Gonguet and Virginie Do and Vish Vogeti and Vítor Albiero and Vladan Petrovic and Weiwei Chu and Wenhan Xiong and Wenyin Fu and Whitney Meers and Xavier Martinet and Xiaodong Wang and Xiaofang Wang and Xiaoqing Ellen Tan and Xide Xia and Xinfeng Xie and Xuchao Jia and Xuewei Wang and Yaelle Goldschlag and Yashesh Gaur and Yasmine Babaei and Yi Wen and Yiwen Song and Yuchen Zhang and Yue Li and Yuning Mao and Zacharie Delpierre Coudert and Zheng Yan and Zhengxing Chen and Zoe Papakipos and Aaditya Singh and Aayushi Srivastava and Abha Jain and Adam Kelsey and Adam Shajnfeld and Adithya Gangidi and Adolfo Victoria and Ahuva Goldstand and Ajay Menon and Ajay Sharma and Alex Boesenberg and Alexei Baevski and Allie Feinstein and Amanda Kallet and Amit Sangani and Amos Teo and Anam Yunus and Andrei Lupu and Andres Alvarado and Andrew Caples and Andrew Gu and Andrew Ho and Andrew Poulton and Andrew Ryan and Ankit Ramchandani and Annie Dong and Annie Franco and Anuj Goyal and Aparajita Saraf and Arkabandhu Chowdhury and Ashley Gabriel and Ashwin Bharambe and Assaf Eisenman and Azadeh Yazdan and Beau James and Ben Maurer and Benjamin Leonhardi and Bernie Huang and Beth Loyd and Beto De Paola and Bhargavi Paranjape and Bing Liu and Bo Wu and Boyu Ni and Braden Hancock and Bram Wasti and Brandon Spence and Brani Stojkovic and Brian Gamido and Britt Montalvo and Carl Parker and Carly Burton and Catalina Mejia and Ce Liu and Changhan Wang and Changkyu Kim and Chao Zhou and Chester Hu and Ching-Hsiang Chu and Chris Cai and Chris Tindal and Christoph Feichtenhofer and Cynthia Gao and Damon Civin and Dana Beaty and Daniel Kreymer and Daniel Li and David Adkins and David Xu and Davide Testuggine and Delia David and Devi Parikh and Diana Liskovich and Didem Foss and Dingkang Wang and Duc Le and Dustin Holland and Edward Dowling and Eissa Jamil and Elaine Montgomery and Eleonora Presani and Emily Hahn and Emily Wood and Eric-Tuan Le and Erik Brinkman and Esteban Arcaute and Evan Dunbar and Evan Smothers and Fei Sun and Felix Kreuk and Feng Tian and Filippos Kokkinos and Firat Ozgenel and Francesco Caggioni and Frank Kanayet and Frank Seide and Gabriela Medina Florez and Gabriella Schwarz and Gada Badeer and Georgia Swee and Gil Halpern and Grant Herman and Grigory Sizov and Guangyi and Zhang and Guna Lakshminarayanan and Hakan Inan and Hamid Shojanazeri and Han Zou and Hannah Wang and Hanwen Zha and Haroun Habeeb and Harrison Rudolph and Helen Suk and Henry Aspegren and Hunter Goldman and Hongyuan Zhan and Ibrahim Damlaj and Igor Molybog and Igor Tufanov and Ilias Leontiadis and Irina-Elena Veliche and Itai Gat and Jake Weissman and James Geboski and James Kohli and Janice Lam and Japhet Asher and Jean-Baptiste Gaya and Jeff Marcus and Jeff Tang and Jennifer Chan and Jenny Zhen and Jeremy Reizenstein and Jeremy Teboul and Jessica Zhong and Jian Jin and Jingyi Yang and Joe Cummings and Jon Carvill and Jon Shepard and Jonathan McPhie and Jonathan Torres and Josh Ginsburg and Junjie Wang and Kai Wu and Kam Hou U and Karan Saxena and Kartikay Khandelwal and Katayoun Zand and Kathy Matosich and Kaushik Veeraraghavan and Kelly Michelena and Keqian Li and Kiran Jagadeesh and Kun Huang and Kunal Chawla and Kyle Huang and Lailin Chen and Lakshya Garg and Lavender A and Leandro Silva and Lee Bell and Lei Zhang and Liangpeng Guo and Licheng Yu and Liron Moshkovich and Luca Wehrstedt and Madian Khabsa and Manav Avalani and Manish Bhatt and Martynas Mankus and Matan Hasson and Matthew Lennie and Matthias Reso and Maxim Groshev and Maxim Naumov and Maya Lathi and Meghan Keneally and Miao Liu and Michael L. Seltzer and Michal Valko and Michelle Restrepo and Mihir Patel and Mik Vyatskov and Mikayel Samvelyan and Mike Clark and Mike Macey and Mike Wang and Miquel Jubert Hermoso and Mo Metanat and Mohammad Rastegari and Munish Bansal and Nandhini Santhanam and Natascha Parks and Natasha White and Navyata Bawa and Nayan Singhal and Nick Egebo and Nicolas Usunier and Nikhil Mehta and Nikolay Pavlovich Laptev and Ning Dong and Norman Cheng and Oleg Chernoguz and Olivia Hart and Omkar Salpekar and Ozlem Kalinli and Parkin Kent and Parth Parekh and Paul Saab and Pavan Balaji and Pedro Rittner and Philip Bontrager and Pierre Roux and Piotr Dollar and Polina Zvyagina and Prashant Ratanchandani and Pritish Yuvraj and Qian Liang and Rachad Alao and Rachel Rodriguez and Rafi Ayub and Raghotham Murthy and Raghu Nayani and Rahul Mitra and Rangaprabhu Parthasarathy and Raymond Li and Rebekkah Hogan and Robin Battey and Rocky Wang and Russ Howes and Ruty Rinott and Sachin Mehta and Sachin Siby and Sai Jayesh Bondu and Samyak Datta and Sara Chugh and Sara Hunt and Sargun Dhillon and Sasha Sidorov and Satadru Pan and Saurabh Mahajan and Saurabh Verma and Seiji Yamamoto and Sharadh Ramaswamy and Shaun Lindsay and Shaun Lindsay and Sheng Feng and Shenghao Lin and Shengxin Cindy Zha and Shishir Patil and Shiva Shankar and Shuqiang Zhang and Shuqiang Zhang and Sinong Wang and Sneha Agarwal and Soji Sajuyigbe and Soumith Chintala and Stephanie Max and Stephen Chen and Steve Kehoe and Steve Satterfield and Sudarshan Govindaprasad and Sumit Gupta and Summer Deng and Sungmin Cho and Sunny Virk and Suraj Subramanian and Sy Choudhury and Sydney Goldman and Tal Remez and Tamar Glaser and Tamara Best and Thilo Koehler and Thomas Robinson and Tianhe Li and Tianjun Zhang and Tim Matthews and Timothy Chou and Tzook Shaked and Varun Vontimitta and Victoria Ajayi and Victoria Montanez and Vijai Mohan and Vinay Satish Kumar and Vishal Mangla and Vlad Ionescu and Vlad Poenaru and Vlad Tiberiu Mihailescu and Vladimir Ivanov and Wei Li and Wenchen Wang and Wenwen Jiang and Wes Bouaziz and Will Constable and Xiaocheng Tang and Xiaojian Wu and Xiaolan Wang and Xilun Wu and Xinbo Gao and Yaniv Kleinman and Yanjun Chen and Ye Hu and Ye Jia and Ye Qi and Yenda Li and Yilin Zhang and Ying Zhang and Yossi Adi and Youngjin Nam and Yu and Wang and Yu Zhao and Yuchen Hao and Yundi Qian and Yunlu Li and Yuzi He and Zach Rait and Zachary DeVito and Zef Rosnbrick and Zhaoduo Wen and Zhenyu Yang and Zhiwei Zhao and Zhiyu Ma},
      year={2024},
      eprint={2407.21783},
      archivePrefix={arXiv},
      primaryClass={cs.AI},
      url={https://arxiv.org/abs/2407.21783}, 
}

@misc{liao2022humancenteredexplainableaixai,
      title={Human-Centered Explainable AI (XAI): From Algorithms to User Experiences}, 
      author={Q. Vera Liao and Kush R. Varshney},
      year={2022},
      eprint={2110.10790},
      archivePrefix={arXiv},
      primaryClass={cs.AI},
      url={https://arxiv.org/abs/2110.10790}, 
}

@inproceedings{liao2020questioning,
author = {Liao, Q. Vera and Gruen, Daniel and Miller, Sarah},
title = {Questioning the AI: Informing Design Practices for Explainable AI User Experiences},
year = {2020},
isbn = {9781450367080},
publisher = {Association for Computing Machinery},
address = {New York, NY, USA},
url = {https://doi.org/10.1145/3313831.3376590},
doi = {10.1145/3313831.3376590},
booktitle = {Proceedings of the 2020 CHI Conference on Human Factors in Computing Systems},
pages = {1–15},
numpages = {15},
location = {Honolulu, HI, USA},
series = {CHI '20}
}

@article{miller_explanation_2019,
	title = {Explanation in artificial intelligence: {Insights} from the social sciences},
	volume = {267},
	issn = {0004-3702},
	shorttitle = {Explanation in artificial intelligence},
	url = {https://www.sciencedirect.com/science/article/pii/S0004370218305988},
	doi = {10.1016/j.artint.2018.07.007},
	language = {en},
	urldate = {2022-02-05},
	journal = {Artificial Intelligence},
	author = {Miller, Tim},
	month = feb,
	year = {2019},
	pages = {1--38},
}

@article{bobrow1977,
title = {GUS, a frame-driven dialog system},
journal = {Artificial Intelligence},
volume = {8},
year = {1977},
issn = {0004-3702},
doi = {10.1016/0004-3702(77)90018-2},
url = {https://www.sciencedirect.com/science/article/pii/0004370277900182},
author = {Daniel G. Bobrow and Ronald M. Kaplan and Martin Kay and Donald A. Norman and Henry Thompson and Terry Winograd},
}

@inproceedings{mitchell2019model,
  title={Model cards for model reporting},
  author={Mitchell, Margaret and Wu, Simone and Zaldivar, Andrew and Barnes, Parker and Vasserman, Lucy and Hutchinson, Ben and Spitzer, Elena and Raji, Inioluwa Deborah and Gebru, Timnit},
  booktitle={Proceedings of the conference on fairness, accountability, and transparency},
  pages={220--229},
  year={2019}
}

@inproceedings{pushkarna2022data,
  title={Data cards: Purposeful and transparent dataset documentation for responsible ai},
  author={Pushkarna, Mahima and Zaldivar, Andrew and Kjartansson, Oddur},
  booktitle={Proceedings of the 2022 ACM Conference on Fairness, Accountability, and Transparency},
  pages={1776--1826},
  year={2022}
}

@article{Rastogi_Zang_Sunkara_Gupta_Khaitan_2020, title={Towards Scalable Multi-Domain Conversational Agents: The Schema-Guided Dialogue Dataset}, volume={34}, url={https://ojs.aaai.org/index.php/AAAI/article/view/6394}, DOI={10.1609/aaai.v34i05.6394}, abstractNote={&lt;p&gt;Virtual assistants such as Google Assistant, Alexa and Siri provide a conversational interface to a large number of services and APIs spanning multiple domains. Such systems need to support an ever-increasing number of services with possibly overlapping functionality. Furthermore, some of these services have little to no training data available. Existing public datasets for task-oriented dialogue do not sufficiently capture these challenges since they cover few domains and assume a single static ontology per domain. In this work, we introduce the the Schema-Guided Dialogue (SGD) dataset, containing over 16k multi-domain conversations spanning 16 domains. Our dataset exceeds the existing task-oriented dialogue corpora in scale, while also highlighting the challenges associated with building large-scale virtual assistants. It provides a challenging testbed for a number of tasks including language understanding, slot filling, dialogue state tracking and response generation. Along the same lines, we present a schema-guided paradigm for task-oriented dialogue, in which predictions are made over a dynamic set of intents and slots, provided as input, using their natural language descriptions. This allows a single dialogue system to easily support a large number of services and facilitates simple integration of new services without requiring additional training data. Building upon the proposed paradigm, we release a model for dialogue state tracking capable of zero-shot generalization to new APIs, while remaining competitive in the regular setting.&lt;/p&gt;}, number={05}, journal={Proceedings of the AAAI Conference on Artificial Intelligence}, author={Rastogi, Abhinav and Zang, Xiaoxue and Sunkara, Srinivas and Gupta, Raghav and Khaitan, Pranav}, year={2020}, month={Apr.}, pages={8689-8696} }

@article{huang2025survey,
  title={A survey on hallucination in large language models: Principles, taxonomy, challenges, and open questions},
  author={Huang, Lei and Yu, Weijiang and Ma, Weitao and Zhong, Weihong and Feng, Zhangyin and Wang, Haotian and Chen, Qianglong and Peng, Weihua and Feng, Xiaocheng and Qin, Bing and others},
  journal={ACM Transactions on Information Systems},
  volume={43},
  number={2},
  pages={1--55},
  year={2025},
  publisher={ACM New York, NY}
}

@inproceedings{tamber-etal-2025-benchmarking,
    title = "Benchmarking {LLM} Faithfulness in {RAG} with Evolving Leaderboards",
    author = "Tamber, Manveer Singh  and
      Bao, Forrest Sheng  and
      Xu, Chenyu  and
      Luo, Ge  and
      Kazi, Suleman  and
      Bae, Minseok  and
      Li, Miaoran  and
      Mendelevitch, Ofer  and
      Qu, Renyi  and
      Lin, Jimmy",
    editor = "Potdar, Saloni  and
      Rojas-Barahona, Lina  and
      Montella, Sebastien",
    booktitle = "Proceedings of the 2025 Conference on Empirical Methods in Natural Language Processing: Industry Track",
    month = nov,
    year = "2025",
    address = "Suzhou (China)",
    publisher = "Association for Computational Linguistics",
    url = "https://aclanthology.org/2025.emnlp-industry.54/",
    doi = "10.18653/v1/2025.emnlp-industry.54",
    pages = "799--811",
    ISBN = "979-8-89176-333-3"
}

@inproceedings{xu-etal-2025-context,
    title = "Does Context Matter? {C}ontextual{J}udge{B}ench for Evaluating {LLM}-based Judges in Contextual Settings",
    author = "Xu, Austin  and
      Bansal, Srijan  and
      Ming, Yifei  and
      Yavuz, Semih  and
      Joty, Shafiq",
    editor = "Che, Wanxiang  and
      Nabende, Joyce  and
      Shutova, Ekaterina  and
      Pilehvar, Mohammad Taher",
    booktitle = "Proceedings of the 63rd Annual Meeting of the Association for Computational Linguistics (Volume 1: Long Papers)",
    month = jul,
    year = "2025",
    address = "Vienna, Austria",
    publisher = "Association for Computational Linguistics",
    url = "https://aclanthology.org/2025.acl-long.470/",
    doi = "10.18653/v1/2025.acl-long.470",
    pages = "9541--9564",
    ISBN = "979-8-89176-251-0"
}

@article{adlakha-etal-2024-evaluating,
    title = "Evaluating Correctness and Faithfulness of Instruction-Following Models for Question Answering",
    author = "Adlakha, Vaibhav  and
      BehnamGhader, Parishad  and
      Lu, Xing Han  and
      Meade, Nicholas  and
      Reddy, Siva",
    journal = "Transactions of the Association for Computational Linguistics",
    volume = "12",
    year = "2024",
    address = "Cambridge, MA",
    publisher = "MIT Press",
    url = "https://aclanthology.org/2024.tacl-1.38/",
    doi = "10.1162/tacl_a_00667",
    pages = "681--699"
}

@inproceedings{kaur2020interpreting,
author = {Kaur, Harmanpreet and Nori, Harsha and Jenkins, Samuel and Caruana, Rich and Wallach, Hanna and Wortman Vaughan, Jennifer},
title = {Interpreting Interpretability: Understanding Data Scientists' Use of Interpretability Tools for Machine Learning},
year = {2020},
isbn = {9781450367080},
publisher = {Association for Computing Machinery},
address = {New York, NY, USA},
url = {https://doi.org/10.1145/3313831.3376219},
doi = {10.1145/3313831.3376219},
booktitle = {Proceedings of the 2020 CHI Conference on Human Factors in Computing Systems},
pages = {1–14},
numpages = {14},
location = {Honolulu, HI, USA},
series = {CHI '20}
}

@inproceedings{peng-etal-2017-composite,
    title = "Composite Task-Completion Dialogue Policy Learning via Hierarchical Deep Reinforcement Learning",
    author = "Peng, Baolin  and
      Li, Xiujun  and
      Li, Lihong  and
      Gao, Jianfeng  and
      Celikyilmaz, Asli  and
      Lee, Sungjin  and
      Wong, Kam-Fai",
    editor = "Palmer, Martha  and
      Hwa, Rebecca  and
      Riedel, Sebastian",
    booktitle = "Proceedings of the 2017 Conference on Empirical Methods in Natural Language Processing",
    month = sep,
    year = "2017",
    address = "Copenhagen, Denmark",
    publisher = "Association for Computational Linguistics",
    url = "https://aclanthology.org/D17-1237/",
    doi = "10.18653/v1/D17-1237",
    pages = "2231--2240"
}

@inproceedings{su-etal-2016-line,
    title = "On-line Active Reward Learning for Policy Optimisation in Spoken Dialogue Systems",
    author = "Su, Pei-Hao  and
      Ga{\v{s}}i{\'c}, Milica  and
      Mrk{\v{s}}i{\'c}, Nikola  and
      Rojas-Barahona, Lina M.  and
      Ultes, Stefan  and
      Vandyke, David  and
      Wen, Tsung-Hsien  and
      Young, Steve",
    editor = "Erk, Katrin  and
      Smith, Noah A.",
    booktitle = "Proceedings of the 54th Annual Meeting of the Association for Computational Linguistics (Volume 1: Long Papers)",
    month = aug,
    year = "2016",
    address = "Berlin, Germany",
    publisher = "Association for Computational Linguistics",
    url = "https://aclanthology.org/P16-1230/",
    doi = "10.18653/v1/P16-1230",
    pages = "2431--2441"
}

@article{borsci2024re,
  title={Re-examining the chatBot Usability Scale (BUS-11) to assess user experience with customer relationship management chatbots},
  author={Borsci, Simone and Schmettow, Martin},
  journal={Personal and Ubiquitous Computing},
  volume={28},
  number={6},
  pages={1033--1044},
  year={2024},
  publisher={Springer}
}

@misc{diabetes_34,
  author       = {Kahn, Michael},
  title        = {{Diabetes}},
   year={1994},
  howpublished = {UCI Machine Learning Repository},
  note         = {{DOI}: https://doi.org/10.24432/C5T59G}
}

@article{lundberg2017unified,
  title={A unified approach to interpreting model predictions},
  author={Lundberg, Scott M and Lee, Su-In},
  journal={Advances in neural information processing systems},
  volume={30},
  year={2017}
}

@inproceedings{van2024xagent,
  title={Xagent: A Conversational XAI Agent Harnessing the Power of Large Language Models},
  author={Nguyen, Van Bach and Schl{\"o}tterer, J{\"o}rg and Seifert, Christin},
  booktitle={Joint Proceedings of the xAI 2024 Late-breaking Work, Demos and Doctoral Consortium co-located with the 2nd World Conference on eXplainable Artificial Intelligence (xAI-2024), Valletta, Malta, July 17},
  number={2024},
  pages={273--280},
  year={2024}
}

@article{
nussbaum2025nomic,
title={Nomic Embed: Training a Reproducible Long Context Text Embedder},
author={Zach Nussbaum and John Xavier Morris and Andriy Mulyar and Brandon Duderstadt},
journal={Transactions on Machine Learning Research},
issn={2835-8856},
year={2025},
url={https://openreview.net/forum?id=IPmzyQSiQE},
note={Reproducibility Certification}
}

@misc{gemmateam2025gemma3technicalreport,
      title={Gemma 3 Technical Report}, 
      author={Gemma Team and Aishwarya Kamath and Johan Ferret and Shreya Pathak and Nino Vieillard and Ramona Merhej and Sarah Perrin and Tatiana Matejovicova and Alexandre Ramé and Morgane Rivière and Louis Rouillard and Thomas Mesnard and Geoffrey Cideron and Jean-bastien Grill and Sabela Ramos and Edouard Yvinec and Michelle Casbon and Etienne Pot and Ivo Penchev and Gaël Liu and Francesco Visin and Kathleen Kenealy and Lucas Beyer and Xiaohai Zhai and Anton Tsitsulin and Robert Busa-Fekete and Alex Feng and Noveen Sachdeva and Benjamin Coleman and Yi Gao and Basil Mustafa and Iain Barr and Emilio Parisotto and David Tian and Matan Eyal and Colin Cherry and Jan-Thorsten Peter and Danila Sinopalnikov and Surya Bhupatiraju and Rishabh Agarwal and Mehran Kazemi and Dan Malkin and Ravin Kumar and David Vilar and Idan Brusilovsky and Jiaming Luo and Andreas Steiner and Abe Friesen and Abhanshu Sharma and Abheesht Sharma and Adi Mayrav Gilady and Adrian Goedeckemeyer and Alaa Saade and Alex Feng and Alexander Kolesnikov and Alexei Bendebury and Alvin Abdagic and Amit Vadi and András György and André Susano Pinto and Anil Das and Ankur Bapna and Antoine Miech and Antoine Yang and Antonia Paterson and Ashish Shenoy and Ayan Chakrabarti and Bilal Piot and Bo Wu and Bobak Shahriari and Bryce Petrini and Charlie Chen and Charline Le Lan and Christopher A. Choquette-Choo and CJ Carey and Cormac Brick and Daniel Deutsch and Danielle Eisenbud and Dee Cattle and Derek Cheng and Dimitris Paparas and Divyashree Shivakumar Sreepathihalli and Doug Reid and Dustin Tran and Dustin Zelle and Eric Noland and Erwin Huizenga and Eugene Kharitonov and Frederick Liu and Gagik Amirkhanyan and Glenn Cameron and Hadi Hashemi and Hanna Klimczak-Plucińska and Harman Singh and Harsh Mehta and Harshal Tushar Lehri and Hussein Hazimeh and Ian Ballantyne and Idan Szpektor and Ivan Nardini and Jean Pouget-Abadie and Jetha Chan and Joe Stanton and John Wieting and Jonathan Lai and Jordi Orbay and Joseph Fernandez and Josh Newlan and Ju-yeong Ji and Jyotinder Singh and Kat Black and Kathy Yu and Kevin Hui and Kiran Vodrahalli and Klaus Greff and Linhai Qiu and Marcella Valentine and Marina Coelho and Marvin Ritter and Matt Hoffman and Matthew Watson and Mayank Chaturvedi and Michael Moynihan and Min Ma and Nabila Babar and Natasha Noy and Nathan Byrd and Nick Roy and Nikola Momchev and Nilay Chauhan and Noveen Sachdeva and Oskar Bunyan and Pankil Botarda and Paul Caron and Paul Kishan Rubenstein and Phil Culliton and Philipp Schmid and Pier Giuseppe Sessa and Pingmei Xu and Piotr Stanczyk and Pouya Tafti and Rakesh Shivanna and Renjie Wu and Renke Pan and Reza Rokni and Rob Willoughby and Rohith Vallu and Ryan Mullins and Sammy Jerome and Sara Smoot and Sertan Girgin and Shariq Iqbal and Shashir Reddy and Shruti Sheth and Siim Põder and Sijal Bhatnagar and Sindhu Raghuram Panyam and Sivan Eiger and Susan Zhang and Tianqi Liu and Trevor Yacovone and Tyler Liechty and Uday Kalra and Utku Evci and Vedant Misra and Vincent Roseberry and Vlad Feinberg and Vlad Kolesnikov and Woohyun Han and Woosuk Kwon and Xi Chen and Yinlam Chow and Yuvein Zhu and Zichuan Wei and Zoltan Egyed and Victor Cotruta and Minh Giang and Phoebe Kirk and Anand Rao and Kat Black and Nabila Babar and Jessica Lo and Erica Moreira and Luiz Gustavo Martins and Omar Sanseviero and Lucas Gonzalez and Zach Gleicher and Tris Warkentin and Vahab Mirrokni and Evan Senter and Eli Collins and Joelle Barral and Zoubin Ghahramani and Raia Hadsell and Yossi Matias and D. Sculley and Slav Petrov and Noah Fiedel and Noam Shazeer and Oriol Vinyals and Jeff Dean and Demis Hassabis and Koray Kavukcuoglu and Clement Farabet and Elena Buchatskaya and Jean-Baptiste Alayrac and Rohan Anil and Dmitry and Lepikhin and Sebastian Borgeaud and Olivier Bachem and Armand Joulin and Alek Andreev and Cassidy Hardin and Robert Dadashi and Léonard Hussenot},
      year={2025},
      eprint={2503.19786},
      archivePrefix={arXiv},
      primaryClass={cs.CL},
      url={https://arxiv.org/abs/2503.19786}, 
}

@misc{touvron2023llama2openfoundation,
      title={Llama 2: Open Foundation and Fine-Tuned Chat Models}, 
      author={Hugo Touvron and Louis Martin and Kevin Stone and Peter Albert and Amjad Almahairi and Yasmine Babaei and Nikolay Bashlykov and Soumya Batra and Prajjwal Bhargava and Shruti Bhosale and Dan Bikel and Lukas Blecher and Cristian Canton Ferrer and Moya Chen and Guillem Cucurull and David Esiobu and Jude Fernandes and Jeremy Fu and Wenyin Fu and Brian Fuller and Cynthia Gao and Vedanuj Goswami and Naman Goyal and Anthony Hartshorn and Saghar Hosseini and Rui Hou and Hakan Inan and Marcin Kardas and Viktor Kerkez and Madian Khabsa and Isabel Kloumann and Artem Korenev and Punit Singh Koura and Marie-Anne Lachaux and Thibaut Lavril and Jenya Lee and Diana Liskovich and Yinghai Lu and Yuning Mao and Xavier Martinet and Todor Mihaylov and Pushkar Mishra and Igor Molybog and Yixin Nie and Andrew Poulton and Jeremy Reizenstein and Rashi Rungta and Kalyan Saladi and Alan Schelten and Ruan Silva and Eric Michael Smith and Ranjan Subramanian and Xiaoqing Ellen Tan and Binh Tang and Ross Taylor and Adina Williams and Jian Xiang Kuan and Puxin Xu and Zheng Yan and Iliyan Zarov and Yuchen Zhang and Angela Fan and Melanie Kambadur and Sharan Narang and Aurelien Rodriguez and Robert Stojnic and Sergey Edunov and Thomas Scialom},
      year={2023},
      eprint={2307.09288},
      archivePrefix={arXiv},
      primaryClass={cs.CL},
      url={https://arxiv.org/abs/2307.09288}, 
}

@article{10.1145/3583558,
author = {Nauta, Meike and Trienes, Jan and Pathak, Shreyasi and Nguyen, Elisa and Peters, Michelle and Schmitt, Yasmin and Schl\"{o}tterer, J\"{o}rg and van Keulen, Maurice and Seifert, Christin},
title = {From Anecdotal Evidence to Quantitative Evaluation Methods: A Systematic Review on Evaluating Explainable AI},
year = {2023},
issue_date = {December 2023},
publisher = {Association for Computing Machinery},
address = {New York, NY, USA},
volume = {55},
number = {13s},
issn = {0360-0300},
url = {https://doi.org/10.1145/3583558},
doi = {10.1145/3583558},
journal = {ACM Comput. Surv.},
month = jul,
articleno = {295},
numpages = {42}
}

@inproceedings{mothilal2020explaining,
  title={Explaining machine learning classifiers through diverse counterfactual explanations},
  author={Mothilal, Ramaravind K and Sharma, Amit and Tan, Chenhao},
  booktitle={Proceedings of the 2020 conference on fairness, accountability, and transparency},
  pages={607--617},
  year={2020}
}

@inproceedings{ross2021explaining,
  title={Explaining NLP models via minimal contrastive editing (MiCE)},
  author={Ross, Alexis and Marasovi{\'c}, Ana and Peters, Matthew E},
  booktitle={Findings of the Association for Computational Linguistics: ACL-IJCNLP 2021},
  pages={3840--3852},
  year={2021}
}

@inproceedings{li-etal-2024-large-language-models,
    title = "Large Language Models as Zero-shot Dialogue State Tracker through Function Calling",
    author = "Li, Zekun  and
      Chen, Zhiyu Zoey  and
      Ross, Mike  and
      Huber, Patrick  and
      Moon, Seungwhan  and
      Lin, Zhaojiang  and
      Dong, Luna  and
      Sagar, Adithya  and
      Yan, Xifeng  and
      Crook, Paul A.",
    editor = "Ku, Lun-Wei  and
      Martins, Andre  and
      Srikumar, Vivek",
    booktitle = "Proceedings of the 62nd Annual Meeting of the Association for Computational Linguistics (Volume 1: Long Papers)",
    month = aug,
    year = "2024",
    address = "Bangkok, Thailand",
    publisher = "Association for Computational Linguistics",
    url = "https://aclanthology.org/2024.acl-long.471/",
    doi = "10.18653/v1/2024.acl-long.471",
    pages = "8688--8704"
}

\appendix

\section{Parsing}
\label{sec:parsing}

\begin{figure*}[t!]
\centering
\resizebox{\textwidth}{!}{
\begin{minipage}{\columnwidth}
\includegraphics[width=\columnwidth]{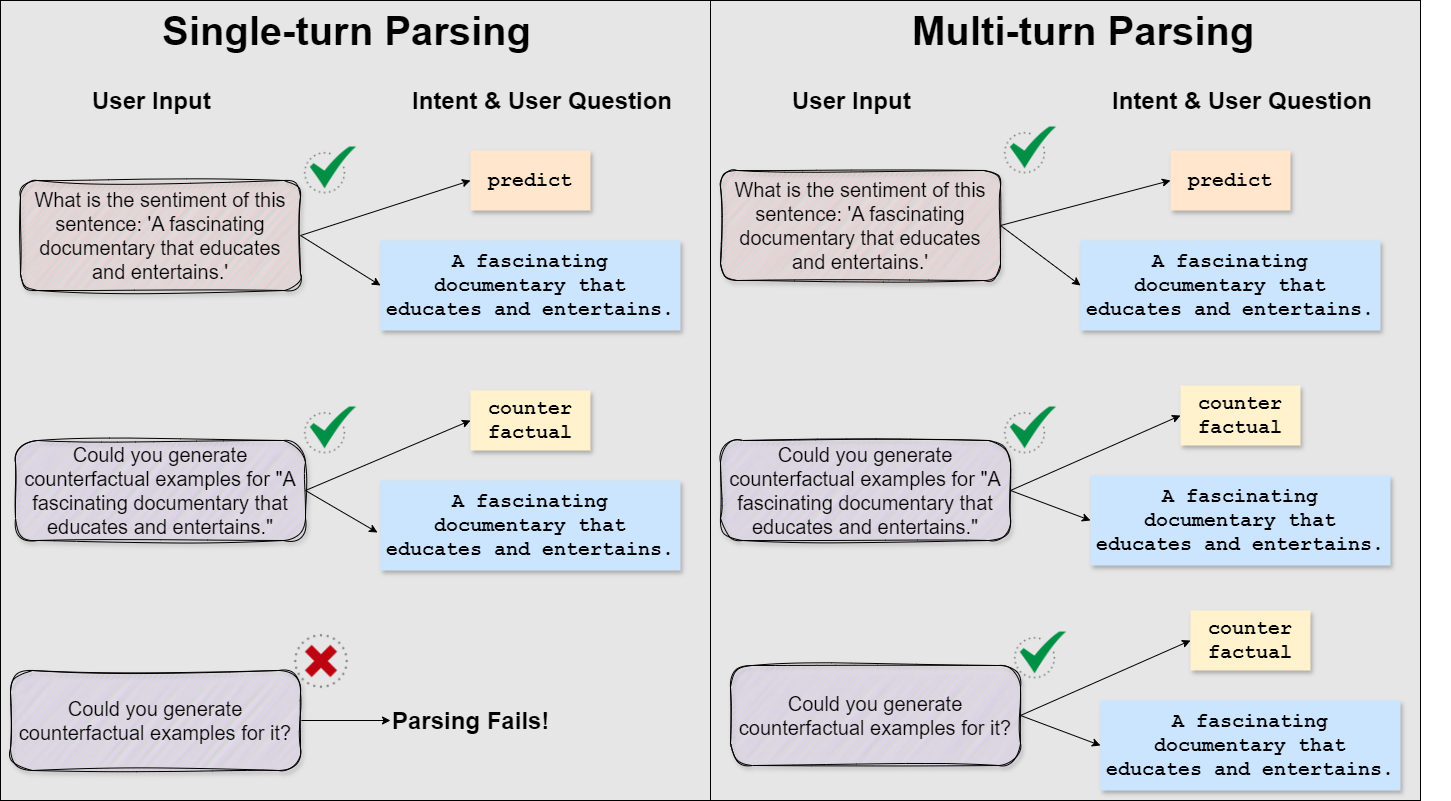}
\end{minipage}
}
\caption{Comparison between single-turn and multi-turn parsing.}
\label{fig:parsing}
\end{figure*}

\paragraph{Failure in single-turn parsing.} Figure~\ref{fig:parsing} shows the comparison between single-turn and multi-turn parsing. For multi-turn parsing, LLMs should be able to utilize the context from previous turns. As illustrated in the third example in Figure~\ref{fig:parsing}, users can simply use pronouns like ``it'' to reference previous questions without repeating the actual question from the last turn. Such cases would fail in single-turn parsing, as there is no conversational context to resolve what ``it'' refers to.

\paragraph{From single-turn to multi-turn.} Table~\ref{tab:test_set}  To build the multi-turn version (Figure~\ref{fig:parsing}, right side), related consecutive questions that operated on the same input text were grouped together into multi-turn conversation sets. Specifically, questions that originally repeated the full instance text had those explicit instance references replaced with pronouns like ``it'' turning them into natural follow-up turns. This way, the conversation flows naturally: the first turn establishes the target text, and subsequent turns refer back to it implicitly — just as a human would in real dialogue. In total, we created 10 multi-turn dialogues with averaging 4–6 turns each.

\begin{figure}[t!]
\centering
\resizebox{\columnwidth}{!}{
\begin{minipage}{\columnwidth}
\includegraphics[width=\columnwidth]{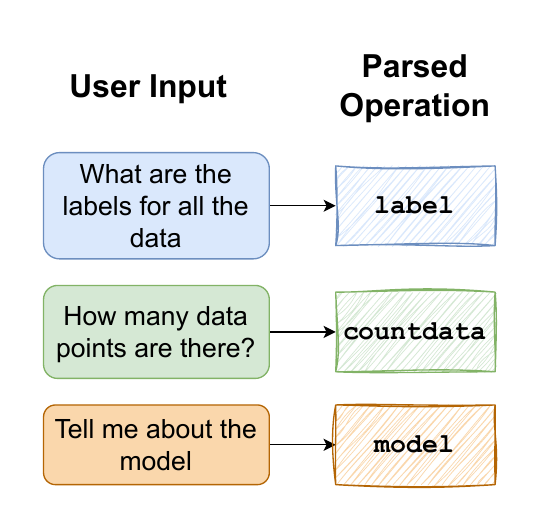}
\end{minipage}
}
\caption{\data{CoXQL} examples.}
\label{fig:coxql}
\end{figure}
\begin{figure}[h]
\centering
\begin{lstlisting}[style=predictionbox]
def explain_how_to_be_that(instance: str, target_label: str = None, number_of_iterations: int = 1) -> str:
    """
    Call this whenever you need to explain how to modify the instance minimally to achieve the target prediction (counterfactual). For example, user asked "How can I change this instance to get prediction X?"
    The tool suggests changes needed to flip the current prediction to the desired one.
    
    Args:
        instance: The input instance.
        target_label: (Optional) The desired prediction or class. If None, suggests changes to flip the current prediction.
        number_of_iterations: Number of counterfactual suggestions to generate.
    
    Output:
        Counterfactual changes (e.g., "increase feature X by 0.2").
    """
    
\end{lstlisting}
\caption{Counterfactual Function Specification Example (Python)}
\label{fig:function-example}
\end{figure}

\section{Realism Validation}
\label{apdx:sec:user_study_realism}
Because the conversations in our dataset were generated using GPT-5.1, we additionally evaluated their realism from a human perspective to verify that they resemble plausible explanatory dialogues.

Specifically, we asked five members of our research group with a background in machine learning to assess the realism of user utterances in the synthetically generated explanatory conversations. Participants were asked to imagine themselves as users seeking to understand how a classification model makes its decisions. They were then presented with generated dialogues between a \textit{User} and an \textit{Explainer}. Their task was to judge whether the User’s messages resembled authentic user questions, reflected a consistent level of knowledge, and remained on topic throughout the conversation.

For each conversation, participants provided a rating on a 5-point Likert scale ranging from 1 (\textbf{completely unrealistic}) to 5 (\textbf{completely realistic}). In addition to the numerical rating, respondents were required to provide a justification for their score, allowing us to collect both quantitative assessments and qualitative rationales regarding perceived conversational realism.

The average score across all conversations was 3.75, suggesting that the generated dialogues achieved a reasonable level of perceived realism.
\section{Qualitative examples}
\label{apdx:sec:qualitative_examples}
In this section, we present additional qualitative results illustrating cases where the original conversational XAI agents fail to capture dialogue context and how \sys{SimXAI} improves their responses. \Cref{apdx:tab:qualitative_results} provides examples for each existing agent. Each agent operates on different tasks and exhibits different types of contextualization failures. \sys{SimXAI} improves these responses across all cases without requiring task-specific or architecture-specific adaptations.

\begin{table*}[t]
\centering
\resizebox{\textwidth}{!}{
\begin{tabular}{c c p{0.4\linewidth} p{0.4\linewidth}}
\toprule
\textbf{Agent} & \textbf{Speaker} & \textbf{Original} & \textbf{\sys{SimXAI}} \\
\midrule

\multirow{6}{*}{TalkToModel}
& User  & \multicolumn{2}{l}{What if the glucose was 50?} \\
& Agent & with the glucose level set to 50, it is still likely to have no diabetes
& with the glucose level set to 50, it is still likely to have no diabetes. \\

& User  & \multicolumn{2}{l}{What if it was 105?} \\
& Agent &  \textcolor{Orange}{... glucose is set to 50.0 and bmi is set to 105.0 is predicted likely to have diabetes.}
& \textcolor{Green}{with the glucose level set to 105, it is still likely to have no diabetes.} \\

\midrule

\multirow{6}{*}{Interrolang}
& User  & \multicolumn{2}{l}{What is the prediction for instance ID 81?} \\
& Agent & The instance with id equal to 81 is predicted True.
& The instance with id 81 is predicted True. \\

& User  & \multicolumn{2}{l}{How to change the instance to get a different prediction?} \\
& Agent & \textcolor{Orange}{This is likely out of my expertise, can I help you with something else?}
& \textcolor{Green}{To get a different prediction, here are the list of words ...} \\
\midrule

\multirow{7}{*}{XAgent}
& User  & \multicolumn{2}{l}{What is the prediction for this instance} \\
& Agent & The prediction for this instance is You will not get the loan....
& The prediction for this instance is you will not get the loan \\

& User  & \multicolumn{2}{l}{Why?} \\
& Agent & \textcolor{Orange}{Sorry, I am not exactly sure whether I understood your question correctly.
Did you mean any of the below?...}
& \textcolor{Green}{The model utilized all features, ... The following graph depicts key features that influence this prediction...} \\

\bottomrule
\end{tabular}
}
\caption{Qualitative results}
\label{apdx:tab:qualitative_results}
\end{table*}
\section{User Study Details}
\label{apdx:sec:user_study_detail}

\subsection{Question Pool}
\label{apdx:subsec:question_pool}
\Cref{apdx:tab:user_study_questions} presents the three tasks and the corresponding question templates that participants were asked to answer. The questions cover several categories of the XAI question taxonomy~\cite{liao2020questioning}, including data understanding, local explanations, and counterfactual inspection.

Some question templates contain placeholders that are replaced with specific values during the study. For these questions, the feature and value are randomly selected when generating the question. The study uses two IDs, which are randomly assigned to the conversational agents.

Each task includes multiple possible questions, and all three tasks are presented to each participant. For each task, questions are randomly selected from the pool without replacement, ensuring that within a single session participants receive as many different questions as possible.

Several selection rules are applied. \textbf{Bolded questions} are always asked, and all types are always included. In addition, \textit{italicized questions} represent follow-up questions that are always selected when their preceding question appears.

\begin{table*}[h]
\centering
\small
\resizebox{\textwidth}{!}{
\begin{tabular}{p{3cm} p{1.5cm} p{8.5cm}}
\hline
\textbf{Task} & \textbf{Type} & \textbf{Question Template} \\
\hline

\multirow{6}{3cm}{Explore the dataset and answer the following question(s).} &
\multirow{6}{1.5cm}{Data Exploration (global)} &
\begin{itemize}[leftmargin=*, nosep]
\item Number of features in the dataset is:
\item Is the feature \tpl{feature} included in the dataset?
\item The maximum value of \tpl{feature} recorded in the dataset is:
\item Find the minimum value of \tpl{feature} recorded in the dataset and choose the correct value:
\end{itemize}
\\
\midrule
\multirow{6}{3cm}{Explore \textbf{patient with ID \tpl{patient\_id}} and answer the following question(s).} &
\multirow{6}{1.5cm}{Data Exploration (local)} &
\begin{itemize}[leftmargin=*, nosep]
\item Find the value of \tpl{feature\_A} for this patient and choose the correct value.
\item What is the value of \tpl{feature\_A} for this patient:
\item \textit{How about the \tpl{feature\_B} value of this patient?}
\item \textit{How about this patient's \tpl{feature\_B} value?}

\end{itemize}
\\
\midrule
\multirow{13}{3cm}{Find the prediction of the model for \textbf{patient with ID \tpl{patient\_id}} and answer the following question(s).} &
\multirow{6}{1.5cm}{Local Explanation} &
\begin{itemize}[leftmargin=*, nosep]
\item \textbf{What does the model predict for this patient?}
\item The most important feature for the prediction is:
\item The least important feature for the prediction is:
\item Is \tpl{feature} used for the prediction for this patient?
\item Top 3 most important features for the prediction are:
\end{itemize}
\\

 &
\multirow{6}{1.5cm}{Counterfactual} &
\begin{itemize}[leftmargin=*, nosep]
\item What would be the prediction if the value of \tpl{feature} is increased to \tpl{value}?
\item What would be the prediction if the value of \tpl{feature} is decreased to \tpl{value}?
\item \textit{What would be the prediction if it is decreased/increased to \tpl{value}?}
\end{itemize}
\\

\hline
\end{tabular}
}
\caption{Tasks and corresponding questions used in the user study.}
\label{apdx:tab:user_study_questions}
\end{table*}

\subsection{Instructions and User Interface}
\begin{figure*}[tbhp]
    \centering
    \begin{subfigure}{0.48\textwidth}
    \includegraphics[width=\linewidth]{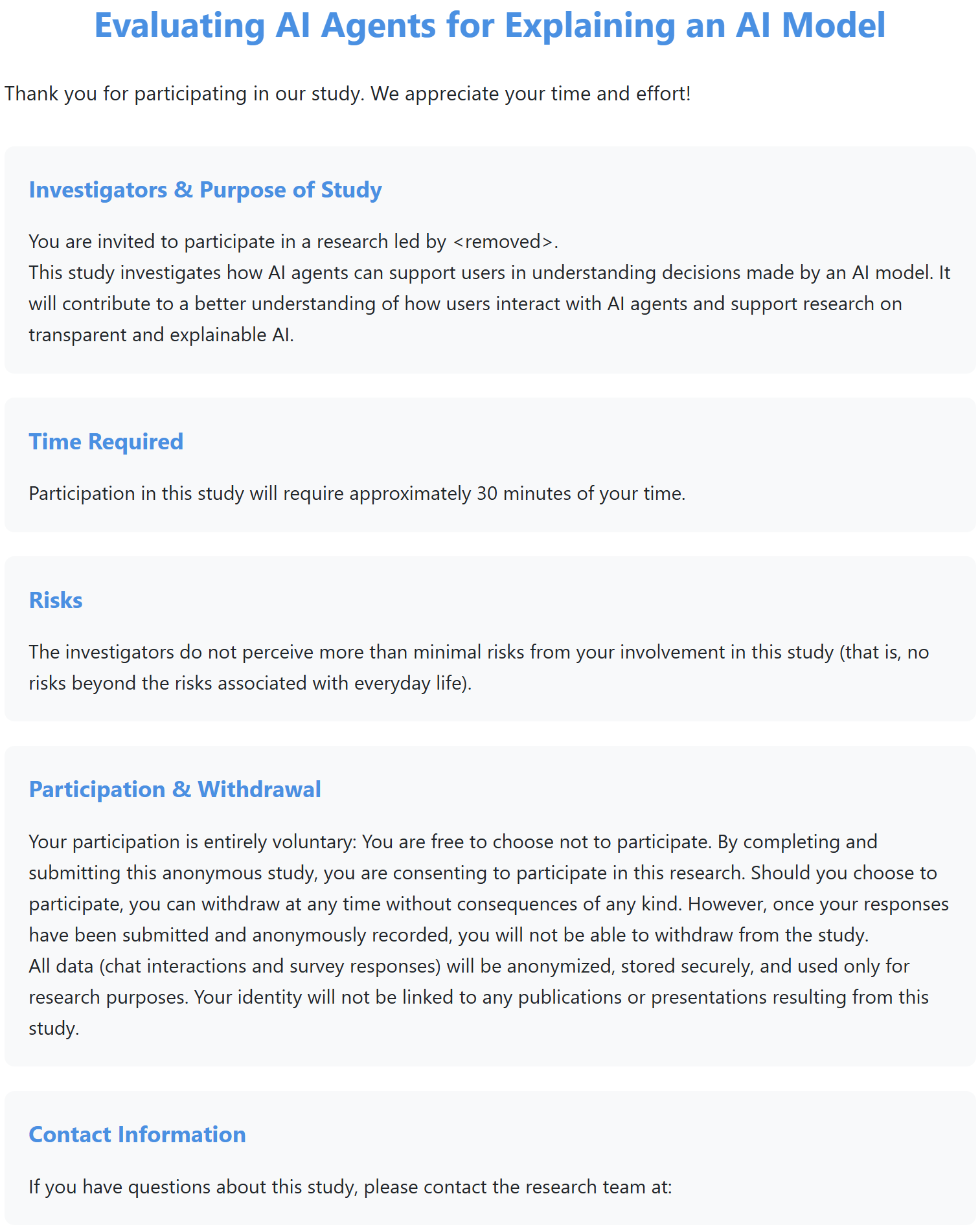}
    \caption{General Instructions.}
    \end{subfigure}
    \begin{subfigure}{0.48\textwidth}
                \includegraphics[width=\linewidth]{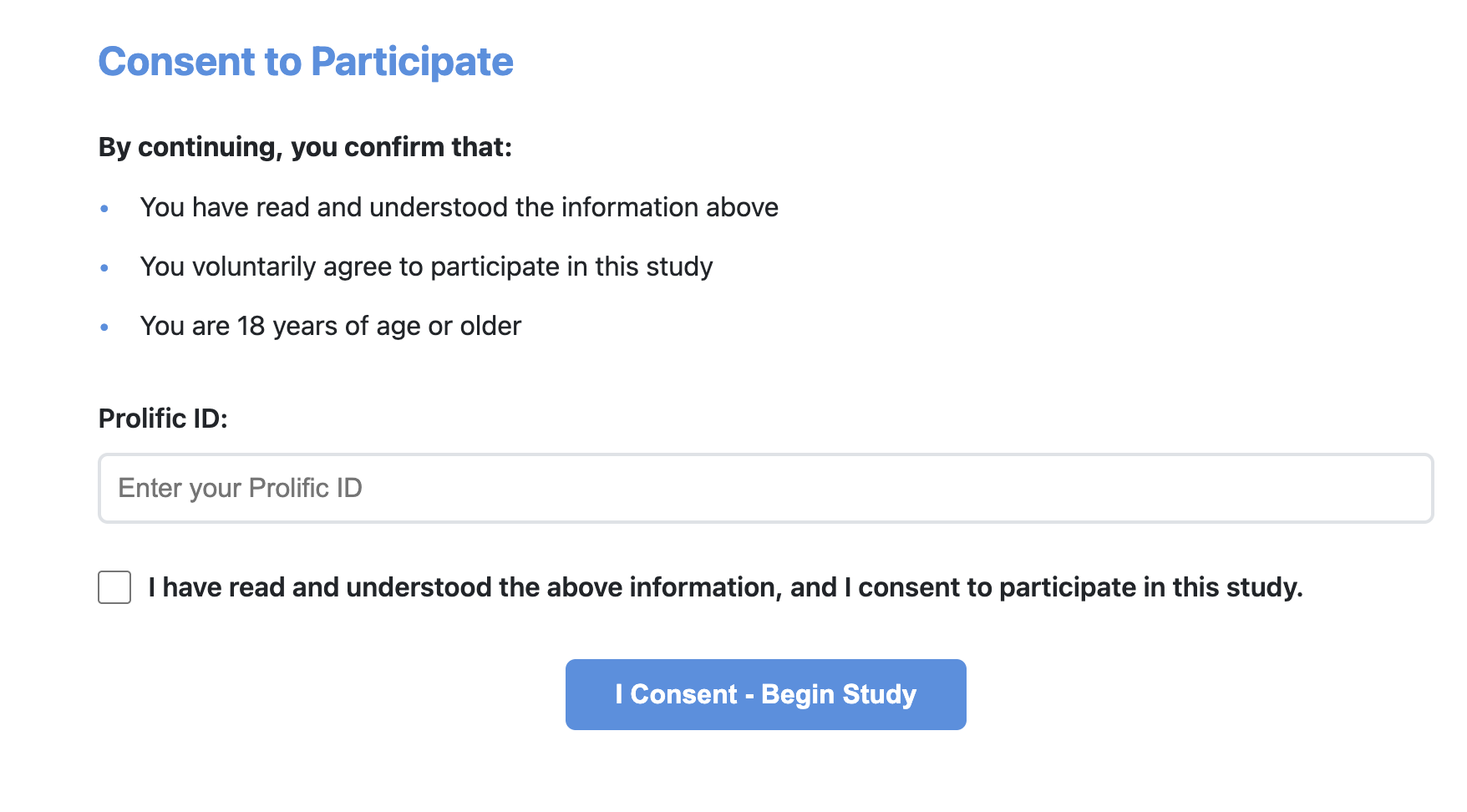}
                \caption{Consent form.}
    \end{subfigure}
    \caption{General instructions and consent form (identifying information removed for blind submission).}
    \label{fig:ux-general-consent}
    
\end{figure*}

\begin{figure*}[t]
    \centering
    \begin{subfigure}{0.48\textwidth}
    \includegraphics[width=\linewidth]{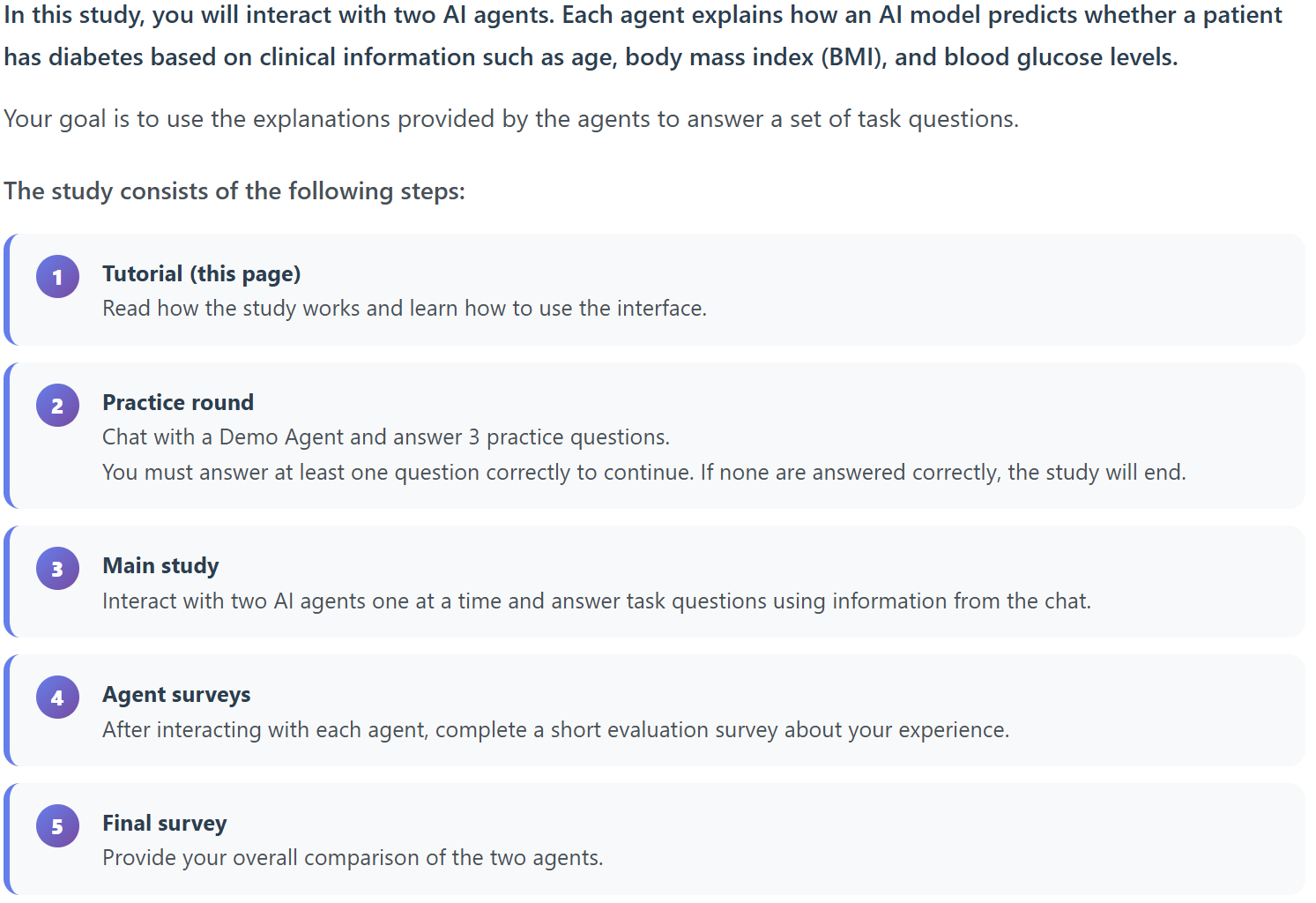}
    \caption{Study overview.}
    \end{subfigure}
    \begin{subfigure}{0.48\textwidth}
                \includegraphics[width=\linewidth]{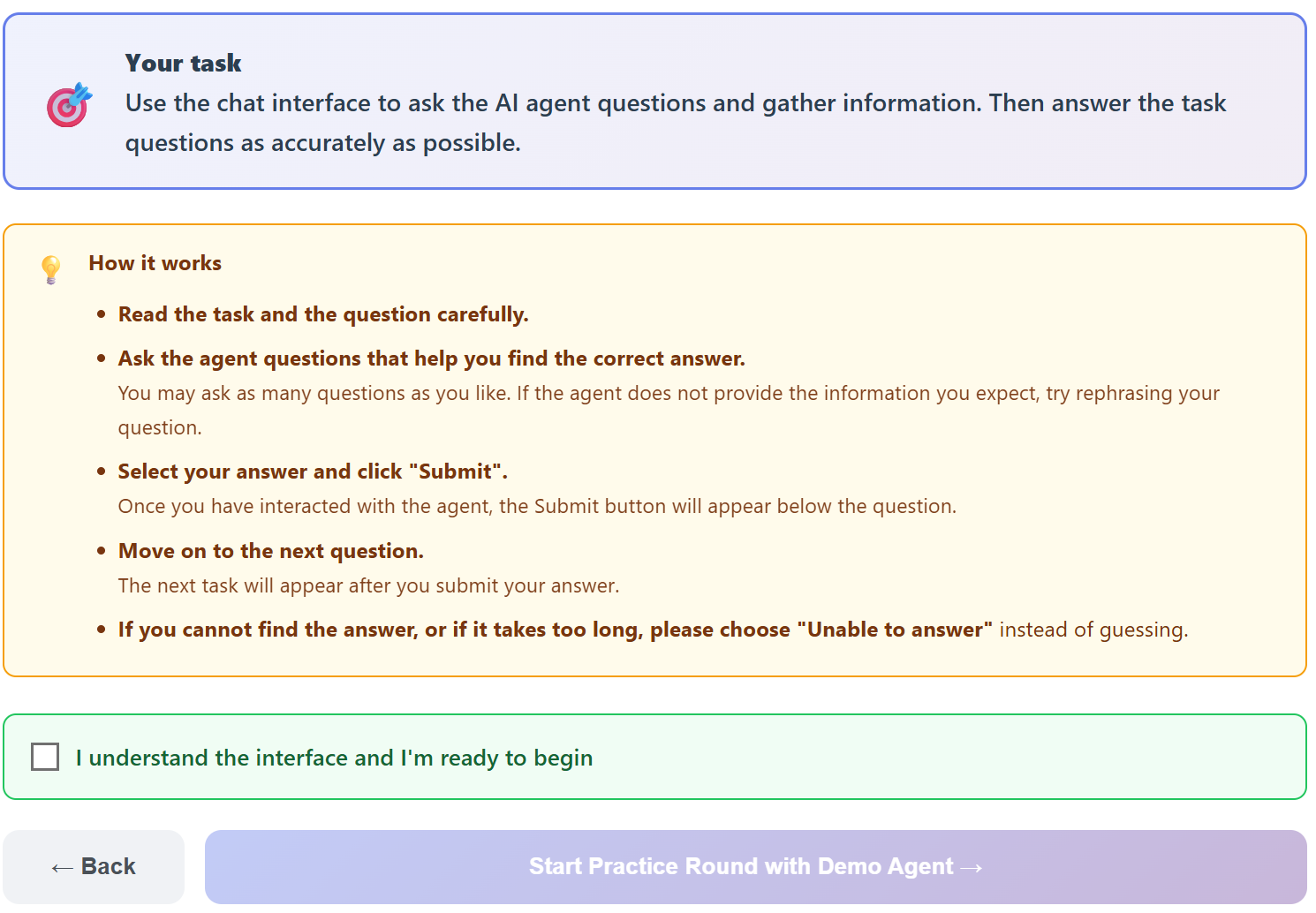}
                \caption{Task instructions.}
    \end{subfigure}
    \caption{Information on the study procedure and how to conduct tasks.}
    \label{fig:ux-overview-task}
\end{figure*}

\begin{figure*}
    \centering
    \includegraphics[width=0.7\textwidth]{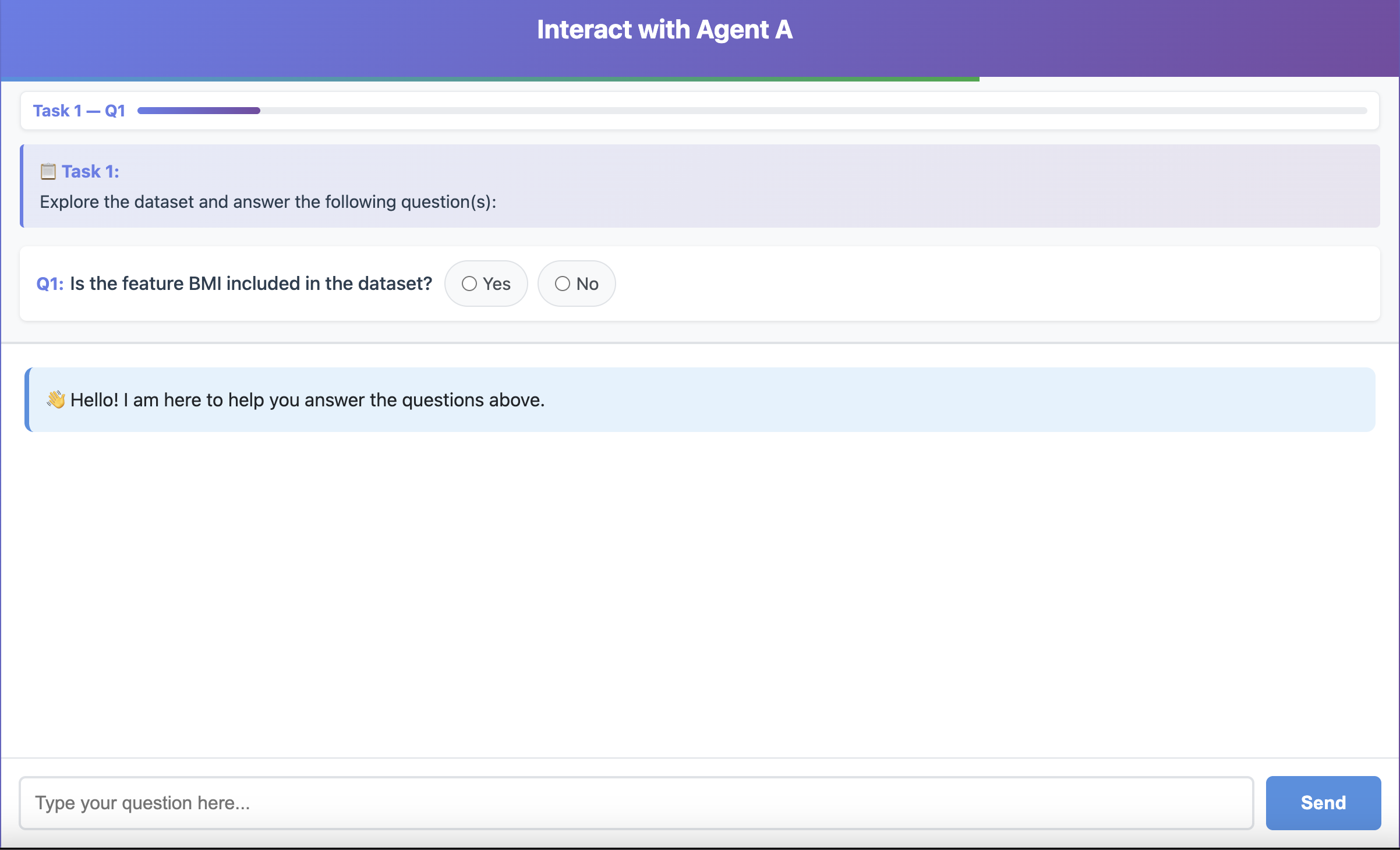}
    \caption{User interface for conversational XAI agent interaction and question answering.}
    \label{fig:ux-interface}
\end{figure*}

\Cref{fig:ux-general-consent} shows screenshots of general instructions and informed consent, \Cref{fig:ux-overview-task} screenshots of information on the overall study procedure and how to conduct the tasks in the study and \Cref{fig:ux-interface} shows the interface, participants interact with.

\subsection{Results}
\label{apdx:subsec:detail_result}
Across 30 completed sessions, integrating \sys{SimXAI} into \sys{TalkToModel} improved both objective task performance and subjective user experience. Unless otherwise noted, we report two-tailed paired $t$-tests over participant-level scores. Table~\ref{tab:userstudy_objective} summarizes the objective outcomes, Table~\ref{tab:userstudy_bus11_means} reports the BUS-11 item means, Table~\ref{tab:userstudy_tests} presents the statistical comparisons, and Table~\ref{tab:userstudy_preference} summarizes the pairwise preferences.

\paragraph{Objective performance.}
As shown in Table~\ref{tab:userstudy_objective}, \sys{SimXAI} substantially improves users' ability to answer explanation-related questions correctly. 
If we count all cases when users indicated that the question is not answerable (``unable to answer'') as incorrect, accuracy increases from 53.3 $\pm$ 19.6 with \sys{TalkToModel} to 84.1 $\pm$ 17.7 with \sys{SimXAI}. The same pattern holds when excluding \textit{unable to answer} responses from the denominator, with accuracy increasing from 59.0 $\pm$ 21.3 to 92.1 $\pm$ 11.6. The corresponding paired tests in Table~\ref{tab:userstudy_tests} confirm that both improvements are highly significant ($p<.001$ in both cases).

The aggregate counts in Table~\ref{tab:userstudy_objective} show the same trend: participants produced many more correct answers with \sys{SimXAI} (227) than with \sys{TalkToModel} (138). Importantly, both systems generated the same number of explicit \textit{unable to answer} responses (24 each), suggesting that the gain came primarily from better answers rather than fewer abstentions.

SimXAI also made interactions more efficient. Compared with \sys{TalkToModel}, it reduced average chat time (7.4 $\pm$ 3.3 minutes vs.\ 5.8 $\pm$ 3.0 minutes) and average conversation length (12.4 $\pm$ 4.1 turns vs.\ 10.0 $\pm$ 1.8 turns). This pattern is also reflected in the turns-per-correct-answer measure reported in Table~\ref{tab:userstudy_objective}, which was notably lower for \sys{SimXAI} (1.32) than for \sys{TalkToModel} (2.70). Overall, users reached correct answers more quickly and with fewer conversational steps.

\paragraph{Subjective evaluation.}
The subjective ratings mirror the objective results. Table~\ref{tab:userstudy_bus11_means} shows that \sys{SimXAI} received higher mean ratings on 10 of the 11 BUS-11 items, and Table~\ref{tab:userstudy_tests} shows that 9 of these differences are statistically significant. The strongest improvements are observed on dimensions central to multi-turn explanatory dialogue, especially clarity, context retention, helpfulness, and informativeness. For example, users rate \sys{SimXAI} higher for clear communication (4.60 vs.\ 3.33), keeping track of conversational context (4.37 vs.\ 3.13), and helping them achieve their goal (4.37 vs.\ 3.10).

The same overall trend holds across the remaining items: \sys{SimXAI} is rated as easier to understand, more accurate, and easier to use for finding information. Only two BUS-11 items do not differ significantly. First, privacy communication is rated relatively low in both conditions, indicating that neither system communicated privacy-related issues particularly well. Second, perceived waiting/response time slightly favors \sys{TalkToModel} numerically, but the difference is not significant. Taken together, the BUS-11 results show that \sys{SimXAI} mainly improves the quality and usefulness of the interaction. These aspects are central to ConvXAI.

\paragraph{Pairwise comparison.}
The direct comparison results in Table~\ref{tab:userstudy_preference} further support this pattern. A large majority of participants prefer \sys{SimXAI} overall (80.0\%), while only 16.7\% prefer \sys{TalkToModel} and 3.3\% report no preference. The same distribution appears for perceived task effectiveness: 80.0\% judge \sys{SimXAI} more helpful, 16.7\% prefer \sys{TalkToModel}, and 3.3\% consider the two systems equally helpful. These preferences are consistent with both the objective performance results and the BUS-11 ratings.

\paragraph{Summary.}
Overall, the results across Tables~\ref{tab:userstudy_objective}--\ref{tab:userstudy_preference} show a consistent pattern: replacing \sys{TalkToModel}'s conversational component with \sys{SimXAI} improves answer quality, increases task success, reduces interaction cost, and leads to more favorable user judgments. The clearest gains appear on aspects that matter most for explanation-oriented multi-turn interaction, namely clarity, contextual coherence, helpfulness, and efficiency.
\begin{table}[h!]
\centering
\small
\begin{tabular}{lcc}
\toprule
\textbf{Comparison item} &  \textbf{\sys{TalkToModel}} & \textbf{\sys{SimXAI}} \\
\midrule
Overall preference & 5 (16.7\%) & 24 (80.0\%) \\
No preference & \multicolumn{2}{c}{1 (3.3\%)} \\
Task effectiveness & 5 (16.7\%) & 24 (80.0\%) \\
Equally helpful & \multicolumn{2}{c}{1 (3.3\%)} \\
\bottomrule
\end{tabular}
\caption{Pairwise comparison results. Most participants preferred \sys{SimXAI} and judged it more effective.}
\label{tab:userstudy_preference}
\end{table}

\begin{table*}[t]
\centering

\begin{tabular}{lcc}
\toprule
\textbf{Metric} &  \textbf{\sys{TalkToModel}} &  \textbf{\sys{SimXAI}} \\
\midrule
Participants & 30 & 30 \\
Accuracy (unable = wrong) & 53.3 $\pm$ 19.6 & 84.1 $\pm$ 17.7 \\
Accuracy (excluding unable) & 59.0 $\pm$ 21.3 & 92.1 $\pm$ 11.6 \\
Total correct / answered (incl. unable) & 138 / 262 & 227 / 270 \\
Total correct / answered (excl. unable) & 138 / 238 & 227 / 246 \\
Unable-to-answer count & 24 & 24 \\
Questions responded & 265 & 270 \\
Chat time (min) & 7.4 $\pm$ 3.3 & 5.8 $\pm$ 3.0 \\
Chat turns & 12.4 $\pm$ 4.1 & 10.0 $\pm$ 1.8 \\
Turns per correct answer & 2.70 & 1.32 \\
\bottomrule
\end{tabular}
\caption{Objective user-study results. \sys{SimXAI} improves accuracy and reduces interaction cost. Means are reported with standard deviations.}
\label{tab:userstudy_objective}
\end{table*}
\begin{table*}[t]
\centering
\resizebox{\linewidth}{!}{
\begin{tabular}{p{0.50\linewidth}lcc}
\toprule
\textbf{BUS-11 item} & \textbf{Short Form} &  \textbf{\sys{TalkToModel}} &  \textbf{\sys{SimXAI}} \\
\midrule
Q1 Chatbot function was easily detectable & Interaction Clarity & 3.80 $\pm$ 0.76 & 4.33 $\pm$ 0.76 \\
Q2 Easy to find information via the chatbot & Information Sufficiency & 4.13 $\pm$ 0.78 & 4.50 $\pm$ 0.57 \\
Q3 Communicating with the chatbot was clear & Interaction Clarity & 3.33 $\pm$ 1.27 & 4.60 $\pm$ 0.56 \\
Q4 Chatbot kept track of conversation context & Context Awareness & 3.13 $\pm$ 1.31 & 4.37 $\pm$ 0.93 \\
Q5 Responses were easy to understand & Response Comprehensibility & 3.60 $\pm$ 1.16 & 4.57 $\pm$ 0.68 \\
Q6 Chatbot helped achieve my goal & Goal Facilitation & 3.10 $\pm$ 1.21 & 4.37 $\pm$ 0.67 \\
Q7 Appropriate amount of information provided & Information Sufficiency & 3.23 $\pm$ 1.17 & 4.37 $\pm$ 0.72 \\
Q8 Only gave needed information & Information Relevance & 3.13 $\pm$ 1.31 & 4.13 $\pm$ 0.94 \\
Q9 Responses were accurate & Response Accuracy & 3.63 $\pm$ 0.96 & 4.47 $\pm$ 0.57 \\
Q10 Informs about privacy issues & Privacy Transparency & 2.30 $\pm$ 1.02 & 2.57 $\pm$ 1.30 \\
Q11 Short waiting / response time & Response Efficiency & 4.37 $\pm$ 0.61 & 4.03 $\pm$ 1.13 \\
\bottomrule
\end{tabular}
}
\caption{Post-interaction BUS-11 ratings (1--5 Likert). \sys{SimXAI} is rated higher on 10 of 11 items.}
\label{tab:userstudy_bus11_means}
\end{table*}
\begin{table*}[t]
\centering

\begin{tabular}{p{0.46\linewidth}cc}
\toprule
\textbf{Metric} & \textbf{t} & \textbf{p} \\
\midrule
Accuracy incl. unable & -9.646 & $<.001^{***}$ \\
Accuracy excl. unable & -8.293 & $<.001^{***}$ \\
\midrule
Q1 Chatbot function was easily detectable & -2.898 & .0071$^{**}$ \\
Q2 Easy to find information via the chatbot & -2.483 & .0190$^{*}$ \\
Q3 Communicating with the chatbot was clear & -5.091 & $<.001^{***}$ \\
Q4 Chatbot kept track of conversation context & -4.570 & $<.001^{***}$ \\
Q5 Responses were easy to understand & -3.778 & $<.001^{***}$ \\
Q6 Chatbot helped achieve my goal & -4.911 & $<.001^{***}$ \\
Q7 Appropriate amount of information provided & -4.335 & $<.001^{***}$ \\
Q8 Only gave needed information & -3.257 & .0029$^{**}$ \\
Q9 Responses were accurate & -3.878 & $<.001^{***}$ \\
Q10 Informs about privacy issues & -1.765 & .0881 \\
Q11 Short waiting / response time & +1.439 & .1608 \\
\bottomrule
\end{tabular}
\caption{Paired two-tailed $t$-tests.}
\label{tab:userstudy_tests}
\end{table*}

\end{document}